\documentclass[a4paper,fleqn]{cas-dc}

\usepackage[authoryear]{natbib}

\makeatletter
\def\thickhline{%
	\noalign{\ifnum0=`}\fi\hrule \@height \thickarrayrulewidth \futurelet
	\reserved@a\@xthickhline}
\def\@xthickhline{\ifx\reserved@a\thickhline
	\vskip\doublerulesep
	\vskip-\thickarrayrulewidth
	\fi
	\ifnum0=`{\fi}}
\makeatother

\newlength{\thickarrayrulewidth}
\usepackage{epigraph}
\usepackage{longtable}

\usepackage{textcomp}

\usepackage{amssymb}

\usepackage{booktabs}

\usepackage{multirow}

\UseRawInputEncoding
\usepackage{times}
\usepackage{epsfig}
\usepackage{amssymb}
\usepackage[ruled,vlined,linesnumbered]{algorithm2e}
\usepackage{booktabs}
\usepackage{caption}
\usepackage{pgfplots}
\pgfplotsset{compat=1.16}
\usepackage{comment}
\usepackage{array, makecell} 
\usepackage{xcolor}
\usepackage{lettrine}
\usepackage{url} 
\usepackage{lscape}
\usepackage{tabularx}
\usepackage{colortbl}
\usepackage{hhline}
\usepackage{vcell}
\usepackage{longtable}
\usepackage{vcell}
\usepackage{rotating}
\usepackage{pdflscape}
\usepackage{sidecap}
\usepackage{neuralnetwork}
\usepackage[export]{adjustbox} 

\usepackage{acronym}
\acrodef{FCN}[FCN]{Fully Convolutional Network}
\acrodef{GAME}[GAME]{Grid Average Mean Absolute Error}
\acrodef{DL}[DL]{Deep Learning}
\acrodef{DNN}[DNN]{Deep Neural Network}
\acrodef{ML}[ML]{Machine Learning}
\acrodef{CV}[CV]{Computer Vision}
\acrodef{AI}[AI]{Artificial Intelligence}
\acrodef{CNN}[CNN]{Convolutional Neural Network}
\acrodef{JCU}[JCU]{James Cook University}
\acrodef{MAE}[MAE]{Mean Average Error}
\acrodef{MAP}[mAP]{Mean Average Precision}
\acrodef{CA}[CA]{Classification Accuracy}
\acrodef{LCFCN}[LCFCN]{Localisation-based Counting loss Fully Convolutional Network}
\acrodef{RUV}[RUV]{Remote Underwater Video}

\newcommand{\blue}[1]{\textcolor{blue}{#1}}
\usepackage[most]{tcolorbox}
\newtcolorbox[auto counter]{pabox}[2][]{%
colback=green!5!white,colframe=green!75!black,fonttitle=\bfseries,
title=Box~\thetcbcounter: #2,#1}

\usepackage{graphicx}
\graphicspath{{./Graphics/}}
\DeclareGraphicsExtensions{.pdf,.jpeg,.png,.jpg}

\usepackage{amsmath}

\usepackage{array}

\usepackage{url}

\def\tsc#1{\csdef{#1}{\textsc{\lowercase{#1}}\xspace}}
\tsc{WGM}
\tsc{QE}

\begin{document}
\begin{frontmatter}

\author[label1]{Alzayat Saleh}
\ead{alzayat.saleh@my.jcu.edu.au}

\author[label1]{Marcus Sheaves}
\ead{marcus.sheaves@jcu.edu.au}

\author[label1,label2]{Dean Jerry}
\ead{dean.jerry@jcu.edu.au}

\author[label1,label2]{Mostafa Rahimi Azghadi \corref{cor1}}
\ead{mostafa.rahimiazghadi@jcu.edu.au}

\cortext[cor1]{Corresponding author.}
\address[label1]{College of Science and Engineering, James Cook University, 1 James Cook Drive, Townsville, 4811, QLD, Australia}
\address[label2]{ARC Research Hub for Supercharging Tropical Aquaculture through Genetic Solutions, James Cook University, 1 James Cook Drive, Townsville, 4811, QLD, Australia}


\begin{abstract}
Marine ecosystems and their fish habitats are 
becoming increasingly important due to their integral role in providing a valuable food source and conservation outcomes. 
Due to their remote and difficult to access nature, marine environments and fish habitats are often monitored using underwater cameras to record videos and images for understanding fish life and ecology, as well as for preserving the environment. There are currently many permanent underwater camera systems deployed at different places around the globe. In addition, there exists numerous studies that use temporary cameras to survey fish habitats. These cameras generate a massive volume of digital data, which cannot be efficiently analysed by current manual processing methods, which involve a human observer. 
\ac{DL} is a cutting-edge \ac{AI} technology that has demonstrated unprecedented performance in analysing visual data. Despite its application to a myriad of domains, its use in underwater fish habitat monitoring remains under explored.  
In this paper, we provide a tutorial that covers the key concepts of \ac{DL}, which help the reader grasp a high-level understanding of how \ac{DL} works. The tutorial also explains a step-by-step procedure on how \ac{DL} algorithms should be developed for challenging applications such as underwater fish monitoring. In addition, we provide a comprehensive survey of key deep learning techniques for fish habitat monitoring including classification, counting, localization, and segmentation. Furthermore, we survey publicly available underwater fish datasets, and compare various \ac{DL} techniques in the underwater fish monitoring domains. 
We also discuss some challenges and opportunities in the emerging field of deep learning for fish habitat processing.
This paper is written to serve as a tutorial for marine scientists who would like to grasp a high-level understanding of \ac{DL}, develop it for their applications by following our step-by-step tutorial, and see how it is evolving to facilitate their research efforts. At the same time, it is suitable for computer scientists who would like to survey state-of-the-art \ac{DL}-based methodologies for fish habitat monitoring. 
\end{abstract}


\shorttitle{Applications of Deep Learning in Fish Habitat Monitoring: A Tutorial and Survey}    

\shortauthors{Saleh et al}


\title [mode = title]{Applications of Deep Learning in Fish Habitat Monitoring: A Tutorial and Survey}  

\end{frontmatter}

\section{Introduction}\label{secintro}
Proper understanding of our planet and its ecosystems is not possible unless suitable tools are developed to explore and learn about our largest ecosystem, the marine environment. \acf{CV} technology through deployment of its underwater cameras can help us better comprehend and manage remote marine fish habitats. However, due to the sheer volume of their visual data, manual processing is time- and cost-prohibitive, requiring a new radical shift in data analysis, through advanced technologies such as \acf{DL}. 

\ac{DL} is at the frontier of computer vision. Its deep neural network architectures are capable of learning complex mappings from high-dimensional data to interpretable feature representations, hence, \ac{DL} has been successfully applied to various challenging computer vision tasks such as semantic image segmentation \citep{Jing2020,Pathak2015ConstrainedCN,Laradji2021AImages,QiAugmentedSupervision,Chuang2011AutomaticSystems}, visual object detection  \citep{Wang2018d,Villon2016a,Kim2016,Pathak2018b}, and tracking  \citep{Garcia2016,Duan2019,kang2018beyond,Lumauag2019a}. These applications have the potential to radically alter the way we interact with the world through computers. Recently, the applications of \ac{DL} and its underlying \acp{DNN} for  underwater visual processing have received significant attention  \citep{Saleh2020,Laradji2021,Villon2018a,Chuang2016,Nilssen2017,Mandal2018c,Naseer2020,Salman2020,Siddiqui2018a}. 

The main advantage of deep learning is its ability to learn features in different data types, such as underwater fish images, through end-to-end training. 
Training of \acp{DNN} is often thought to be easy. 
Many frameworks take delight in providing few lines of code that solve some \ac{CV} tasks, providing the misleading impression that all that is needed is then plug and play, using some general Application Programming Interfaces (APIs). 
In these APIs, the developers have lifted the burden from us and, in doing so, disguised the complexity behind a few lines of code needed to achieve the task at hand.
The framework developers have achieved the purpose of "providing a few lines of code" but we, the end-users, have been fooled to believe we need to spend only a few hours learning the intricacies of the provided APIs. 

However, when it comes to training a DL algorithm, things become more complicated. The task of training a \ac{DNN} is actually as complicated as the problem it is intended to solve. In fish monitoring for example, the number of input images you use, how you pre-process your images, how you build your models, how you fine-tune the model (using dropout or regularization, for example), how you extract the features, how you combine them to produce final predictions, what metric you use to report your model performance, and your choice of which layer to extract features from to feed to your classifier, are among some of the many variables to consider when training a DNN. You can  include any number of variations on these factors to further optimize your model and to achieve the best possible accuracy.

Due to the above intricacies, most of the time \acp{DNN} are not simply an "off-the-shelf" technology that works with all kind of datasets, even those similar to the one that has been meticulously customised for it. 
The fact that training a customised high-performance \ac{DNN} is rigorous and challenging is now widely accepted.
However, this challenging process can be facilitated by being patient, paying attention to details, and working systematically.
Developing customised \ac{DNN}s with a specific application, for example, for underwater fish monitoring, should follow the same systematic steps of developing any other computer vision applications ( e.g. detection of vehicles in traffic). The only difference lies in the type of data being fed to the DNN. 

In this paper, we first present a tutorial that covers the background of DL to help understand the above-mentioned common \ac{DL} terminologies. The tutorial also provides a comprehensive overview of the essential systematic steps 
to help better develop a supervised \ac{DL} model, with a focus on underwater fish habitat monitoring. 

In the second part of the paper, we survey state-of-the-art research and development on the use of \ac{DL} for fish monitoring. We synthesize the literature into four main categories covering the common \ac{CV} tasks of classification, counting, localization, and segmentation of fish images. We investigate different deep learning architectures and their performance.
We also survey publicly available underwater fish image datasets. 
Finally, we provide a comprehensive overview of the challenges in applying \ac{DL} to marine fish monitoring domains. We also draw a roadmap for future research works. 

Although a number of previous relevant review articles  \citep{Goodwin2021,Li2021a,Zhao2021,Yang2021,li2021tracking,Moniruzzaman2017,saleh2022a} exist, our paper has a different approach and motivation that compliments prior surveys. 
Compared to  \citep{Goodwin2021}, which provides a survey of the general domain of ecological data analysis, covering a wide array of studies on plankton, fish, marine mammals, pollution, and nutrient cycling, we focus only on fish monitoring. We also provide a detailed analysis of fish datasets and comprehensively review the literature on four key tasks in underwater fish video and image processing. This detailed analysis and review is not provided in  \citep{Goodwin2021}, or any of the previous works, making our paper useful for readers who would like to study fish monitoring using DL in more details and depth, while seeing a comprehensive literature review. 

In addition,  \citep{Li2021a} provides a review of studies on fish condition, growth, and behavior monitoring in aquaculture settings. It briefly covers and reviews various DL architectures and their aquaculture applications, unlike the present communication that is focused mainly on \ac{CNN} and provides a detailed survey and analysis of the underwater fish monitoring literature.

The work presented in  \citep{Zhao2021} covers the general domain of Machine Learning, as opposed to the specific domain of DL in our paper. This is done for aquaculture applications as wide as fish biomass and behavior analysis to water quality predictions, while also briefly covering and reviewing fish classification and detection methods. 

A survey of computer vision models for fish detection and behavior analysis in digital aquaculture is provided in  \citep{Yang2021}.  
An interested reader should study  \citep{Yang2021} before reading our paper, due to the background technical details provided on image acquisition, which are key to developing effective DL datasets and models, as we discussed in our paper.

Furthermore, the DL-based studies presented in  \citep{li2021tracking} and  \citep{Moniruzzaman2017} are mainly around the two specific tasks of underwater fish tracking, and underwater object detection, respectively. These applications are different to our study. However, since our underwater fish monitoring task are related to these applications, our paper can complement these works. 

In  \citep{saleh2022a}, we have provided a historical survey of fish classification methods between the years 2003-2021. These methods cover traditional CV techniques and modern DL methods, only for fish classification in underwater habitats and not for the general domain of underwater fish habitat monitoring.

\section{Deep Learning}\label{secdl} 

Deep learning is a sub-field of machine learning composed of interrelated algorithms and concepts used in training a deep neural network  \citep{saleh2022a}. One of the main reasons behind the extereme popularity of deep learning is the unprecedented and unparalleled performance it has achieved across different fields especially image recognition.

Deep learning utilizes multi-layered neural networks for automatic learning of input features. Features are distinguishing properties of learning inputs e.g. the color or shape of different fish. The deep learning concept was first proposed based on the idea that the traditional multi-layer artificial neural networks, could learn complex nonlinear features and their relations with more generalization and at a rapid speed. 
To learn deep features efficiently, researchers found that a modified version of neural networks, i.e. \ac{CNN}, works very well in the image processing field  \citep{saleh2022a}. 
In the following sections, we will first introduce the basic concepts of neural networks in general and then describe CNNs and explain how they learn and then process input images.

\subsection{Neural Networks}

A 'neural network' is a computational model that is inspired by biological neural systems and uses simple, non-linear, computational rules to mimic these systems.  Neural networks are composed of simple processing elements called, neurons. 
By organising neurons in a layered structure, interconnecting them and changing the weights associated with each interconnection, a 'neural network' can be trained to solve a complex problem, such as recognising if a fish is present in an image. It is then possible to store the connections between neurons for later use. Training a neural network to perform different tasks e.g. recognizing fish in an image, or determining where a fish is in an underwater image, is called the 'learning process'. During supervised learning (explained later), the inputs to the network are presented with each input having a desired output. The learning process determines which interconnections (weights) are most important to the system for learning the task at hand and mapping all the inputs to all their desired outputs, as best as possible. 

The general idea of neural networks is to have layers of neurons for learning the input data. There are three consecutive layer types in a neural network, i.e. input, hidden, and output. 
The hidden layers can learn the patterns in the data passed to the network through the input layer. It is within the hidden layers that  classification, or in some cases regression, of the input data takes place. The hidden layers can learn abstract patterns and features in the data on their own. In general, there will be more layers in a \ac{DNN} compared to artificial (shallow) networks for image classification tasks, and this is why \acp{DNN} are called deep and can achieve higher accuracies.

\subsubsection{Neuron}
The neuron, also known as a node or perceptron in a neural network, is its basic unit of computing. The neuron takes inputs from other nodes and produces an output. Every input has a weight $w$ that is allocated based on its relative significance to other inputs. 
As depicted in Figure \ref{fig:cell}, the node applies the activation function $f$ (described below) on the weighted sum of its inputs.

\subsubsection{Activation Functions}
The activation function  \citep{Vogels2005} in a neural network defines whether a given node is "activated" or not based on the weighted sum of input features. The sigmoid function is one of the most commonly used activation functions. It is defined as:
\begin{equation}
S(x)=\frac{1}{1+e^{-x}}
\end{equation}
where $S(x)$ is the sigmoid function output that will be used as the input for the following node and $x$ is the weighted sum of input features from the previous layer. The sigmoid function is non-linear and its value ranges between 0 and 1. Sigmoid is popular in image classification because its 0-1 range can be represented as the probability of "activating" each output class. The output with the largest "activation" value is then selected, thus facilitating the network's ability to classify the image.

\subsubsection{Bias Node}
\begin{figure}[!ht]
\centering	
	\begin{tikzpicture}[shorten >=1pt]
  		\tikzstyle{unit}=[draw,shape=circle,minimum size =1.4cm, color=black ,line width=0.01cm]
  		\tikzstyle{org}=[draw,shape=circle,minimum size =2cm, color=blue ,line width=0.03cm]
 
 		\node[unit](km) at (-3,3){$x_1$};
		\node[unit](kn) at (-3,1){$x_2$};
		\node[unit](ko) at (-3,-1){$x_3$};
       	\node[org](i) at (0,1){$f(\sum_{i=1}^{n}{x_iw_i + b})$};
    	\node[unit](k1) at (3,1){$Y$};

		\node at (-1.8,2.50){$w_1$};
		\node at (-1.8,1.25){$w_2$};
		\node at (-1.8,0.2){$w_3$};
 
    	\draw[->] (i) -- (k1);
		\draw[->] (km) -- (i);
		\draw[->] (kn) -- (i);
		\draw[->] (ko) -- (i);
		
    	\end{tikzpicture}
\caption{Diagram of the perceptron neuron. Inputs $x_{i}$ are multiplied in the weights $w_{i}$. The neuron body (\blue{blue}) accumulates the sum of all multiplication inputs and then fires an output signal $Y$ according to its activation function $f$. 
}
\label{fig:cell}
\end{figure}
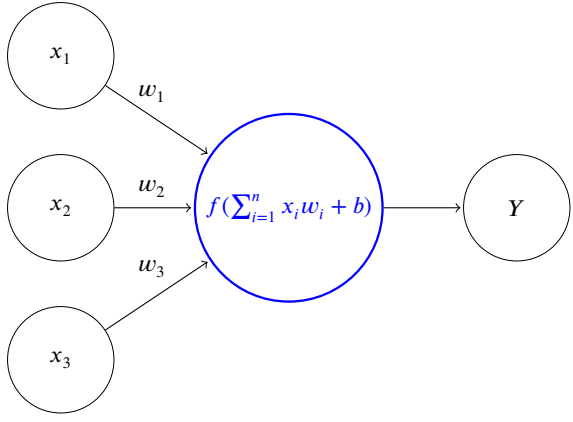

Another important component in successful neural networks are the "bias" nodes, which, as shown in Fig. \ref{fig:cell}, add a bias value $b$ to the sum of input-weight multiplications to increase the model's flexibility. In particular, when all input features equal to 0, the network can adjust to the data and decrease the distance between the fitted values in other data spaces.

\subsubsection{Loss Function}
In machine learning, there is always a function that needs to be decreased or increased to reach the closest possible mapping between the input and output domains. This function is usually known as the objective function. When it needs to be minimised, for instance for the case of neural network supervised learning, we might refer to it as the cost, loss, or error function. Although different \ac{DL} publications may define specific meanings for some of these terms, we use them indiscriminately in this paper.
In general, loss functions measure the performance of a data-based \ac{ML} model. 
The loss function is important to consider, as it measures and presents learning error in the form of a single real number between predicted values and expected values.
As an example, the loss function for linear regression is defined as:

\begin{equation}L=\frac{1}{2 m} \sum_{i=1}^{m}(\hat{y}-y)^{2},\end{equation}
where $m$ is the number of training examples, $\hat{y}$ is the predicted value of the model, and $y$ is the true value of the inputs in the training data.

For classification tasks, such as fish species classification, the loss function $L$ is generally a cross-entropy loss function. \textit{Cross-entropy loss} measures the performance of a classification model with a probability value ranging from 0 to 1. The loss of cross-entropy functions will increase as the predicted probability differs from the ground truth. Another classification loss is Hinge Loss. In \textit{Hinge Loss}, the correct category score should, by some safety margin, be higher than the sum of values for all incorrect categories.

\subsubsection{Optimization}
\label{opt}
In supervised learning, the learning task can be reduced to an optimization problem in the form of 
\begin{equation}
    \theta^{*}=\arg\min _{\theta} g(\theta),
\end{equation}
where $\theta$ is a parameter vector, at which the loss function $g(\theta)$ that usually represent the average loss for all training examples, reaches its minimum. $g$ can be represented as
\begin{equation} \label{eq:loss}
    g(\theta)=\frac{1}{n} \sum_{i=1}^{n} L\left(f_{\theta}\left(x_{i}\right), y_{i}\right),
\end{equation}
where $(x_{i}, y_{i})$ represents a (input, desired output) training pair.

Similarly, in \ac{DL}, an optimization method is used to train the neural network by minimising the error function $E$ that is defined as
\begin{equation}
E(W, b)=\sum_{i=1}^{m} L\left(\hat{y}_{i}, {y}_{i}\right)
\end{equation}
where $W$ and $b$ are the weights and biases of the network, respectively. The value of the error function $E$ is thus the sum of the mean squared loss $L$ between the predicted value $\hat{y}$ and true value $y$, for m training examples. The value of $\hat{y}$ is obtained during the forward propagation step and makes use of the previously-mentioned weights and biases of the network, which can be initialised in different ways. Optimization minimizes the value of the error function $E$ by updating the values of the trainable parameters $W$ and $b$.

The error function $E$ is usually minimised by using its gradient slopes for the parameters. The most commonly used 
optimization
method is \textit{Gradient Descent}  \citep{Sun2019}, in which the gradient is optimized by calculating a matrix of partial derivatives (computed using backpropagation, as detailed in the next subsection). These derivatives provide the slope of $g$ simultaneously at each dimension of $\theta$. 
Therefore, the gradient is used to determine the next direction to search for the Global Optimum. To enhance $\theta$ and reach a lower $g$, a small quantity is subtracted from  $\theta$ in the optimal direction (since the gradient provides the direction of the rise and conversely the descent in $g$), such that the global optimum is eventually reached and $g$ is minimized.

\subsubsection{Backpropagation}
Backpropagation is probably the most important part of learning in neural networks. It is performed after a forward propagation or pass, in which a subset of the training dataset (named a batch) $\left\{\left(x_{i}, y_{i}\right)\right\}_{i=1}^{m}$ and the current network parameters $\theta$ are used to calculate the final layer output and the loss. During the forward pass, the data input is passed to the first layer to process according to its activation function and their values are passed on to the next layer, hence the term "forward pass". 
 After the forward pass and calculating the final layer loss, backpropagation happens, through which we start to calculate the loss backwards, layer by layer, and the layer derivatives are then "chained" by the local gradients to minimise the overall loss, $g$.

\subsubsection{Regularization} Regularization is another important concept in neural networks learning. It is a technique that makes small changes to the learning algorithm to improve the performance of the model on testing or out-of-sample data  \citep{Bisong2019}. In other words, it avoids the risk of over-fitting the training data by discouraging the formation of complex mapping functions or models. Model regularization involves a regularization term being added to the general model loss function, which takes into account the loss function value for all the training dataset examples. 
Thus, when using regularization, the loss function $g(\theta)$ (described in Eq.~\ref{eq:loss}) becomes
\begin{equation}
    g(\theta)=\frac{1}{n} \sum_{i=1}^{n} L\left(f_{\theta}\left(x_{i}\right), y_{i}\right)+R\left(f_{\theta}\right),
\end{equation}
where, $R\left(f_{\theta}\right)$ is the added regularization function.

The most common forms of regularization are L1 and L2  \citep{Ng2004}. The difference between them is that L2 is the sum of the square of the weights, while L1 is the sum of the weights.

\subsection{Convolutional Neural Network (CNN)} \label{sec:cnn}

The most powerful class of \acp{DNN} are convolutional neural networks. As their name infers, convolutional networks work by performing a convolution (filtering) operation on the input data. A CNN is usually composed of several convolution layers, which extract useful features from the input data by sliding  convolution filters across the input image represented to the network as matrices. %
One of the first successful examples of the use of CNNs in computer vision was AlexNet proposed by Krizhevsky \emph{et al.} in 2012. AlexNet achieved great success only using four convolutional layers. 
Since 2012, many different flavours of \acp{CNN} have been proposed using different architectures and count of convolutional and other complementary layers. These architectures have revolutionized computer vision and image processing in different domains from agriculture  \citep{Olsen2019DeepWeeds:Learning} to medicine  \citep{Saleh2021AImages}.
CNNs have also been widely applied in underwater visual monitoring and processing for counting, localizing, classifying, and segmenting objects of interest such as fish  \citep{Saleh2020a}.

\begin{figure*}[htbp]
\centering
\includegraphics[width=0.99\textwidth]{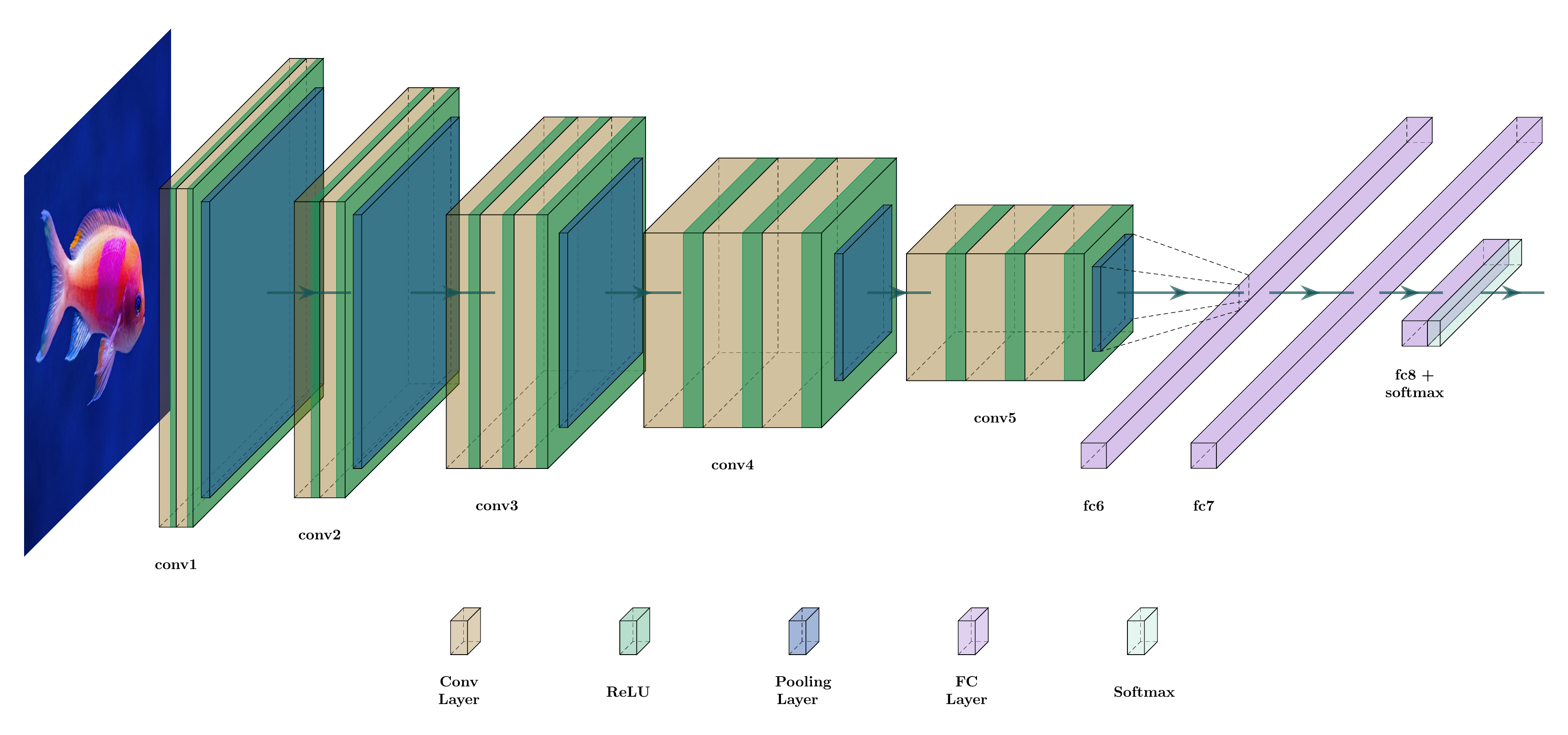}
\caption{Schematic diagram of a CNN architecture used for the classification of fish images. The architecture consists of five convolutional layers that include the batch norm operation within them, followed by pooling layers (conv1-conv5). In this model, the feature maps from convolutional layers are pooled through pooling layers then flattened through two fully connected layers (fc6 and fc7). The classification output is the result of a fully connected layer and a softmax activation layer (fc8+softmax).}
\label{fig:fish_cnn}
\end{figure*}

A typical CNN architecture is composed of convolutional layers, pooling layers, non-linear activation layers, and final output layers, as shown in Figure \ref{fig:fish_cnn}.
It is through the filtering convolution operation combined with other parts of the CNN that useful features of the input data are extracted and learned automatically. The learning of a CNN usually involves finding the appropriate number, size, and structure of convolution filters, pooling layers, and activation functions and their parameters during training and seeing various examples of the inputs. In the below subsections, we will cover these basic building blocks and layers of a typical \ac{CNN}.

\begin{itemize}
   \item \textbf{Convolutional layer}: As already mentioned, a convolutional layer applies a filtering (convolution) operation on its input matrix data to generate another matrix called a feature map. The input matrix can contain the input image information or the feature map generated by a previous CNN layer. 
   The feature maps are the core of a CNN, where useful features of an input are extracted and learned across several convolutional layers. %

    \item \textbf{Batch Normalization}: The goal of this operation, which follows the convolutional operation, is to normalize the learning of the network across the current set of training data (batch), hence the name batch normalization. This is done to improve the speed of learning and the convergence of the deep learning model, because otherwise, the network may see very wide variety of features extracted in its convolutional layers, due to wide input variations. Batch normalization happens by subtracting its input mean and dividing the result by its standard deviation.

    \item \textbf{Activation layer}: This layer that follows the batch normalization layer is the normal neuron activation function explained earlier. It is used to increase the non-linearity of the convolutional layer output and increase its power in learning complex data. The most common activation functions used in conjunction with convolutional layers are Rectified Linear Unit (ReLU) and Sigmoid. Activation functions are also used in the final non-convolutional fully-connected  layers of a CNN. A common output activation function is Softmax.

    \item \textbf{Pooling layer}: The output feature map of the convolutional layer that is batch normalized and passed the activation function, is often too big for the next convolutional layer to handle. To reduce its size and improve the efficiency of computation, it can be pooled in a pooling layer to generate a reduced sized feature map, while keeping important features.
    Pooling is a common operation in CNNs and is used in almost all practical convolutional networks. The most common pooling layers are max pooling and average pooling.

    \item \textbf{Dropout}: To avoid overfitting to the training data, dropout operations is introduced after the pooling layers. Their task is to cut the network's dependence to a single data instance at each traing step, by randomly removing (dropping out) features extracted using the previous convolutional layer.

    \item \textbf{Fully connected layer}: Fully connected layer, also known as dense layer, is the second last layer of a CNN, before the output layer. This layer contains a small number of neurons, each of which connected to every neuron in the previous layer. So the network is said to be fully connected. The fully connected layer takes all the inputs and weights from the previous layer, and combines them together into a single vector or matrix. This vector is then passed through an activation function, such as the sigmoid, to calculate output values of the CNN generated by its final output layer. 
    
\end{itemize}

\subsection{Supervised Learning}\label{secsl}

There are two main approaches to learning in general \ac{DL}. These include unsupervised and supervised learning. Unsupervised learning is often used to discover the structure and composition of the input and output domains without explicit and supervised target domain. This approach enables generalization from one input domain to another by transforming data representations that are not directly related to the data distribution of target domain.

The supervised learning approach, on the other hand, is designed to explicitly map data from the input domain to its output domain via training pairs that exhibit matching representations. These pairs are carefully crafted by a human (supervisor), hence the name.  
The training process of supervised learning can suffer from instability and is less effective than the unsupervised learning method, because it learns with an accurate target distribution without domain-specific knowledge.

Supervised deep learning uses a subtle deep neural network mechanism to extract useful features from large amounts of input training data that are labelled to show their desired output domain. 
The learning is done by using the repetitive backpropagation process  \citep{Rojas1996} explained earlier, to adjust the DL architecture internal parameters, such as the shape, number, and size of convolutional, pooling, and fully connected layers, that have been used to determine the representation in each layer from the representation in the preceding layer. In general, adjusting the DL architecture and its parameters to do the best mapping of the input training data to their desired output, as best as possible, is the same as optimising a function $f$, through backpropagation, to map the input domain $X$, to its matching output domain $Y$, \textit{i.e.} ($f: X \mapsto Y$).

\section{Developing Deep Learning Models}\label{sectech}

\begin{figure}[tbp]
\centering
\includegraphics[width=0.48\textwidth]{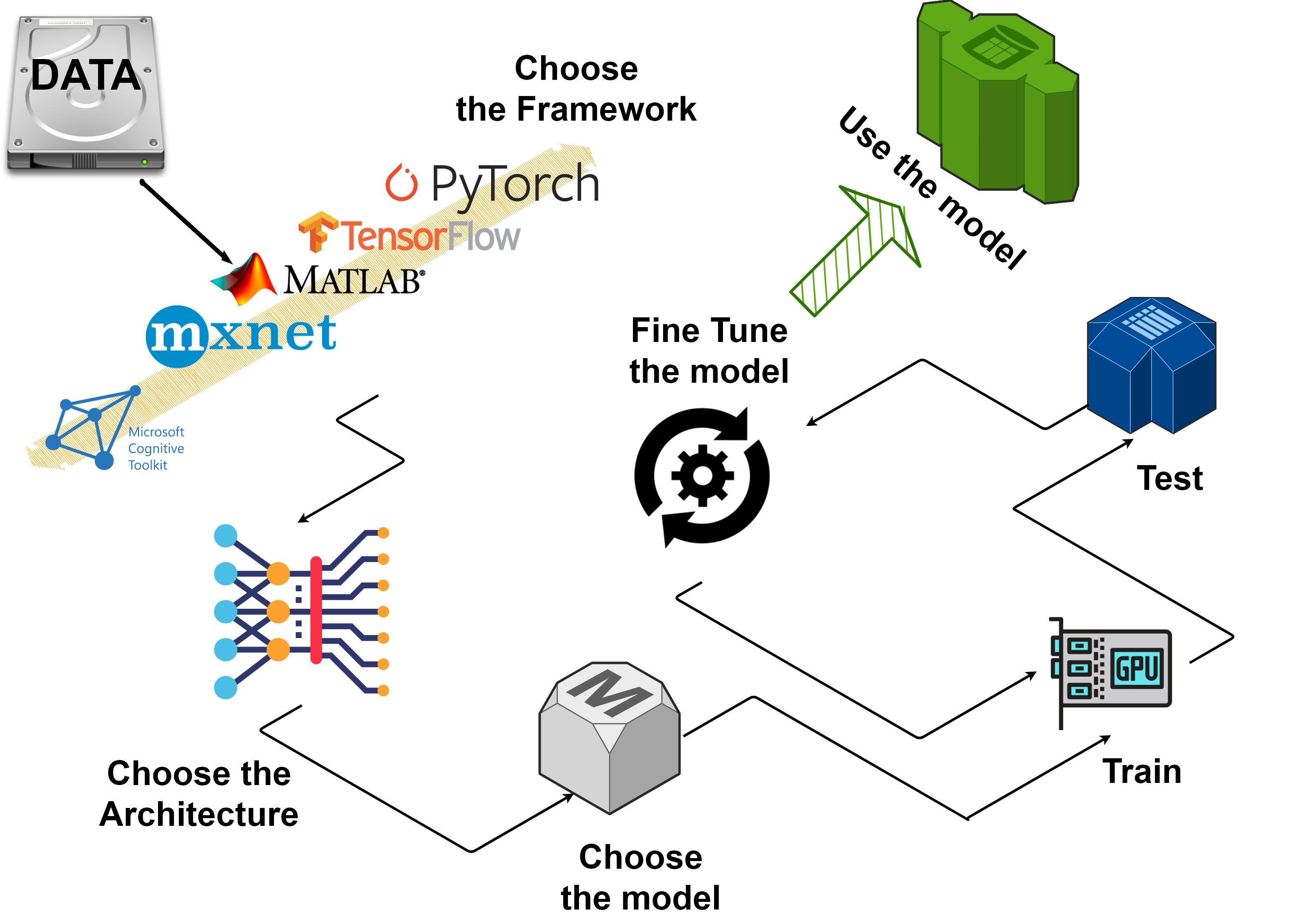}
\caption{A schematic diagram showing the steps and components required for training a deep learning model.}
\label{fig:train}
\end{figure}

A comprehensive overview of the essential systematic steps for training a \ac{DL} model is summarized in Figure \ref{fig:train}. 
Even though these steps are general in \ac{DL} training, we included useful tips arising from our experience in developing \ac{DL} applications in various domains from medical imaging to marine science applications. Nevertheless, we put an emphasis on the development of \ac{DL} for underwater fish habitat monitoring.

\subsection{Training Dataset}
The available training data is essential for developing an efficient \ac{DL} model. Datasets are becoming increasingly crucial, even more so than algorithms. 
Perhaps, the most important factor when considering a supervised learning dataset is its size. 
The requirement for a large training dataset to achieve high accuracy is often a big obstacle. 
Because visual algorithms are trained by pairs of images and labels, in a supervised manner, they can only identify what has already been given to them. 
As a result, depending on the project, the number of objects to identify, and the required performance, training datasets might contain hundreds to millions of images.
However, smaller training datasets with only a few hundred samples per class may also achieve good results  \citep{Saleh2020a,Konovalov2019a,Konovalov2018,Konovalov2019b}.
Nevertheless, the larger the training dataset, the greater the recognition accuracy. 

Because of the scarcity of datasets and the difficulty of acquiring reliable data, approaches for boosting the accuracy rate from small samples will inevitably become a focus of future studies. 
The problem of limited sample data can be also alleviated by transfer learning  \citep{Mathur2020,Molchanov2016,Lee2018}. 
Furthermore, data augmentation will become increasingly critical.
Section \ref{seclimit} covers some challenges of limited data and some approaches to address these challenges.

The second factor to consider when preparing a dataset for \ac{DL} training is having a balance.
This is critical to ensure that each class to be identified contains a sufficient number of instances to minimise class imbalance biases. These biases happen when the \ac{DL} favours one or more classes due to seeing them more often when being trained.

Also, the training dataset is typically divided into two subsets, the training subset for efficiently training the model and the validation/test subset for assessing the trained model's performance.
For the training subset, a subset of the training dataset is reserved for training the model. If the training subset is too large, it can prolong the model training. If, on the other hand, the training subset is too small, the resulting model may not generalise well to unseen inputs. 
The validation/test subset is typically used to avoid overfitting, which is a common problem in machine learning and happens when the developed model simply memorises the inputs rather than properly learning them. 
Cross-validation is another widely used methodology for testing a \ac{DL} model's training performance, by splitting the training dataset into multiple mutually exclusive subsets of training and testing data. One method of cross-validation is called $k-fold$ cross-validation, in which the training dataset is split into $k$ equally sized subsets. In this method,  $k-1$ folds are used for training the model, while the remaining fold is used to test the learning performance. 
This process is repeated until all the folds have been used once as a test/validation set.

In addition to the above, it is usually vital to, initially and before embarking on code development, perform a comprehensive inspection of the dataset. This will help to clean the dataset, for instance by finding and removing duplicate data instances. It also helps identify imbalances and biases, as well as data distribution, trends, or outliers, which will help in better model design and understanding of possible wrong \ac{DNN} predictions.

\begin{figure*}[htbp]
\centering
\includegraphics[width=0.99\textwidth]{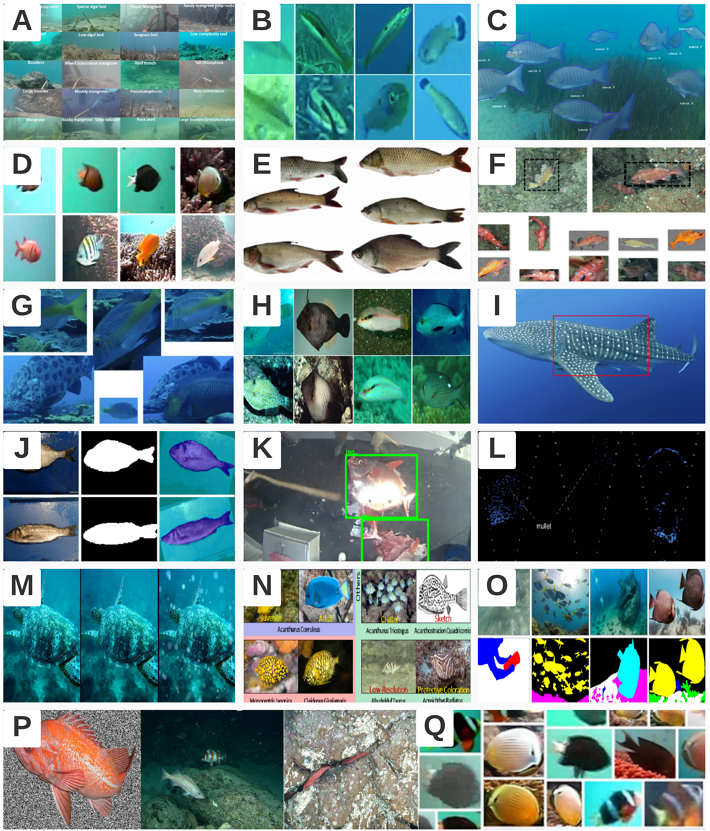}
\caption{Sample images from publicly available datasets detailed in Table \ref{table:dataset}.}
\label{fig:datasets}
\end{figure*}

Fortunately, in the domain of fish habitat monitoring, researchers currently have access to a variety of datasets. 
Table \ref{table:dataset} lists publicly available underwater fish datasets, their sources, and where to get them, in addition to a summary of their features, their labels, and their sizes. The main point to note about these datasets is that they differ in both size and the number of features. 
Although the number of these fish datasets is still small (17), the diversity of aquatic species they cover is already quite wide. They cover a large number of aquatic species, as indicated in Fig. ~\ref{fig:datasets}. Moreover, each dataset features a different number of images that have varying resolutions. For each image, there is also a ground truth annotated by a human expert, which make them very useful. For instance,  these datasets can be used by researchers to test their \ac{DL} models or to pre-train them, as the first step, for their more specific fish monitoring tasks.

After preparing the training dataset or utilising alternative approaches to addressing insufficient data challenge, one can start developing their \ac{DL} model using a machine-learning development framework. 

\subsection{Development framework}
The rapid evolution of \ac{DL} has led to the creation of a vast number of development libraries and packages that enable the setting up of \acp{DNN} with insignificant effort.  
Usability and availability of resources, architectural support, customisability, and hardware support are all various benefits of using existing machine-learning frameworks. 
The most commonly used frameworks are PyTorch, Tensorflow, MATLAB, Microsoft Cognitive Toolkit (CNTK) and Apache MXNET. 
In the context of \ac{DL} for marine research, as will be shown later in Tables \ref{table:cont} to \ref{table:seg}, PyTorch and TensorFlow are the dominant frameworks, while Matlab and Caffe have been used only in a few works. 
Overall, details such as the project needs and the programmer and developer preference should be taken into account, when choosing the development framework. 

When the development framework is chosen, the next step is to find the most suitable network architecture for the task at hand. This sometimes depends on the framework, as some recent methods may not immediately be supported by all frameworks.

\begin{table*}
\begin{sideways}
\begin{minipage}{\textheight}

\centering
\caption{Summary of some publicly available datasets containing fish for training and testing deep learning models.}
\label{table:dataset}
\arrayrulecolor[rgb]{0.647,0.647,0.647}
\resizebox{\linewidth}{!}{%
\begin{tabular}{>{\hspace{0pt}}m{0.154\linewidth}>{\hspace{0pt}}m{0.150\linewidth}>{\hspace{0pt}}m{0.075\linewidth}>{\hspace{0pt}}m{0.244\linewidth}>{\hspace{0pt}}m{0.355\linewidth}} 
\arrayrulecolor{black}\hline
\textbf{ Dataset } & \textbf{ Summary } & \textbf{ Labels } & \textbf{ Dataset size } & \textbf{ Website } \\ 
\arrayrulecolor[rgb]{0.647,0.647,0.647}\hline
\rowcolor[rgb]{0.929,0.929,0.929}A - \textbf{ Deepfish } & Videos
  from coastal habitats in north-eastern and
  western
  Australia & fish/no
  fish
    &  40k
  classification labels,
   3.2k images
  with point-level
  annotations,
  310
  segmentation 
  masks & \href{https://github.com/alzayats/DeepFish}{github.com/alzayats/DeepFish} \\
B - \textbf{ Croatian Fish Dataset } & 12 species of
  fish found in Croatian waters & species names & 794
  classification labels & \href{http://www.inf-cv.uni-jena.de/fine\_grained\_recognition.html\#datasets}{www.inf-cv.uni-jena.de/fine\_grained\_recognition.html\#datasets} \\
\rowcolor[rgb]{0.929,0.929,0.929}C -  \textbf{ Fish in seagrass habitats } & RUV
  taken in Australian
  seagrass
  habitat of 2
  species & species &  9k
  classification labels,
  bounding
  boxes and
  segmentation
  masks & \href{https://github.com/globalwetlands/luderick-seagrass}{github.com/globalwetlands/luderick-seagrass} \\
D - \textbf{ Fish4Knowledge } & Fish
  detection and tracking
  dataset, 17
  videos at 10 min
  long, rate of
  5 fps. & fish/no fish &  3.5k bounding
  boxes & \href{https://groups.inf.ed.ac.uk/f4k/index.html}{groups.inf.ed.ac.uk/f4k/index.html} \\
\rowcolor[rgb]{0.929,0.929,0.929}E - \textbf{ Fish-Pak } & Image
  dataset of 6 different
  fish
  species from 3 locations
  in
  Pakistan & species &  1k classification
  labels & \href{https://data.mendeley.com/datasets/n3ydw29sbz/3}{data.mendeley.com/datasets/n3ydw29sbz/3} \\
F -\textbf{ Labeled Fishes in the Wild } & Rockfish (Sebastes
  spp.) and other species (non-fish)
  near the
  seabed & fish/non-fish &  1k bounding
  boxes (fish),
   3k (non-fish) & \href{https://swfscdata.nmfs.noaa.gov/labeled-fishes-in-the-wild/}{swfscdata.nmfs.noaa.gov/labeled-fishes-in-the-wild/}\\
\rowcolor[rgb]{0.929,0.929,0.929}G - \textbf{ OzFish } & Large
  data set comprising of 507 species of fish. & species,
  fish/no fish &  80k labeled
  cropped images,  45k bounding
  box
  annotations
  (fish/no fish) & \href{https://github.com/open-AIMS/ozfish}{github.com/open-AIMS/ozfish} \\
H - \textbf{ QUT Fish Dataset } & 468 species
  in varying
  ex-situ and in-situ
  habitats. & species name &  4k
  classification images & \href{https://www.dropbox.com/s/e2xya1pzr2tm9xr/QUT\_fish\_data.zip?dl=0}{www.dropbox.com/s/e2xya1pzr2tm9xr/QUT\_fish\_data.zip?dl=0}\\
\rowcolor[rgb]{0.929,0.929,0.929}I - \textbf{ Whale Shark ID } & 543
  individual whale sharks
   (Rhincodon typus) & individuals &  7.8k bounding
  boxes & \href{http://lila.science/datasets/whale-shark-id}{http://lila.science/datasets/whale-shark-id}\\
J - \textbf{Large Scale Fish Dataset } &  9 different seafood types collected from a supermarket in Izmir, Turkey & species name & For each class, there are 1000 augmented images and their pair-wise augmented ground truths& \href{https://www.kaggle.com/crowww/a-large-scale-fish-dataset}{www.kaggle.com/crowww/a-large-scale-fish-dataset} \\
\rowcolor[rgb]{0.929,0.929,0.929}K - \textbf{ NCFM } & Image
  dataset of 8 different
  fish
  species & species
  name & \textasciitilde{}16000 classification images & \href{https://www.kaggle.com/c/the-nature-conservancy-fisheries-monitoring/data}{www.kaggle.com/c/the-nature-conservancy-fisheries-monitoring/data}\\
L - \textbf{ Mugil liza sonar } & Sonar-based
  underwater videos of schools of migratory mullets (Mugil liza) & number of
  fish & 500 counting
  images & \href{https://zenodo.org/record/4751942\#.YKzfUKgzayk}{zenodo.org/record/4751942\#.YKzfUKgzayk} \\
\rowcolor[rgb]{0.929,0.929,0.929}M - \textbf{ MSRB Dataset } & Real
  underwater images without marine snow and synthesized with marine snow & NA & \textasciitilde{}6000
  images & \href{https://github.com/ychtanaka/marine-snow}{github.com/ychtanaka/marine-snow}\\
N - \textbf{ WildFish } & 1,000 fish
  categories & species name & \textasciitilde{}54000 classification
  images & \href{https://github.com/PeiqinZhuang/WildFish}{github.com/PeiqinZhuang/WildFish}\\
\rowcolor[rgb]{0.929,0.929,0.929}O - \textbf{ SUIM } & Image
  dataset of 8 different
  underwater
  objects & object
  name & \textasciitilde{}1500
  annotated images 
  semantic
  segmentation mask & \href{https://github.com/xahidbuffon/SUIM}{github.com/xahidbuffon/SUIM} \\
P - \textbf{DZPeru fish-datasets     } & Several species
  in varying
  ex-situ and in-situ
  habitats. & species name & \textasciitilde{}17000 annotated
  images 
  segmentation
  mask & \href{https://github.com/DZPeru/fish-datasets}{github.com/DZPeru/fish-datasets} \\
\rowcolor[rgb]{0.929,0.929,0.929} Q - \textbf{LifeCLEF} & 10
  different fish species & species
  name & \textasciitilde{}1000
  annotated videos & \href{https://www.imageclef.org/}{www.imageclef.org/} \\
\hline
\end{tabular}
}
\arrayrulecolor{black}

\end{minipage}
\end{sideways}

\end{table*}

\subsection{Network Architecture}
Network architecture is the structure of the \ac{DL} model, which depends on what it intends to achieve and its expected input and output.
Therefore, the type of  training dataset and the expected outcome influence the architecture's choice and its performance.
\ac{DL} network architectures can differ in a variety of ways such as the type and number of layers, their structure, and their order. 
Before selecting a network architecture, it is critical to understand the dataset you have and the task you are going to complete.
For example, convolutional neural networks or \acp{CNN} are known to learn higher-order features, such as colours and shapes, from data within their convolution layers. Therefore, they are ideally adapted in image-based object recognition.
On the other hand, Recurrent Neural Networks (RNNs) have the capability of processing temporal information or sequential data, such as the order of words in a sentence.
This feature is ideal for tasks such as handwriting or speech recognition. 

In the context of fish habitat monitoring, if you are working on a task that requires you to learn temporal information of the input sequence, for example fish image sequence analysis, the \ac{DL} architecture you choose can be very important. For example, a \acp{CNN} architecture is more suited for \emph{image-based} object recognition such as fish classification, while the RNN architecture is more suitable for tasks where the input sequence is temporal in nature such as generating fish habitat descriptions. 

To find a suitable architecture, you first need to define your problem. This problem is defined by two questions: (1) What features will you extract? (2) How will you label these features? The features you extract are defined by your data. In other words, you are interested in the representation of the data you have. 
The number of features you choose to extract is defined by the task you are trying to solve. As described above, the \ac{DL} architectures can learn features such as colours and shapes from image-based object recognition.  Before trying to construct your network, you first need to decide what data type you will use and how will you encode the information. 
After you have defined your task, you should think about what features are important for the task. You will need to define this in order to construct your network. For example, if the features you want to extract are fish shape and fish location, then you could define a convolutional architecture.
The features you choose to define should be a subset of all the features in the data. For example, for an image-based object recognition network, you would extract features such as fish species. However, your extracted features will also need to cover all the data. For example, you will also need features of the type of water or the type of background.  It is important to take all these features into account when defining your network.
For a complete discussion on different \ac{DL} architectures see  \citep{Khan2020a}.

\subsection{Network Model}
When a general network architecture is selected, the next step is to select, or sometimes develop, a network model of that architecture. For instance, when you decided to use a CNN, you can use different varieties of \ac{CNN} models. The rule of thumb for selecting a \ac{CNN} is to choose a model that results in a satisfactory training loss for your dataset. 
Creating an exotic and creative model is not recommended at this stage. It is usually recommended to avoid the temptation and choose a model big enough to overfit your dataset, and then regularise it properly to improve the validation loss.

For example, one may pick a well-known \ac{CNN} model, e.g. ResNet, which can be used out-of-the-box, if their task is simple, e.g. fish classification.
In later stages, they can customise their model to adequately capture their dataset. We show in Tables \ref{table:cont} to \ref{table:seg} in the next section that ResNet is the most commonly used model for fish counting (Table \ref{table:cont}), fish localization (Table \ref{table:loc}), and fish segmentation (Table \ref{table:seg}).

\subsection{Training the model}
After choosing the best model is the time to set up a full train/validation pipeline.
The below steps are recommended at this stage of development.
    \begin{itemize}
        \item Start with a simple model (i.e. a small number of convolutional layers) that can hardly go wrong and visualise the model performance metrics. Do not use an out-of-the-box large model like ResNet, just yet. 
        It is recommended to plot training loss to see how the network is progressing during learning and if the loss is getting smaller. This also shows the speed of learning. 
        \item To better understand the process, it is recommended to use a fixed random seed (for randomly initialising the network parameters) to ensure that the same results can be achieved when running the code twice. 
        \item Do not perform any data augmentation at this stage as it may introduce errors. You can do data augmentation at a later stage after confirming that your network works properly. You can see a brief introduction to data augmentation and other methods at subsection \ref{secgen}.
        \item Use ADAM algorithm  \citep{Kingma2014Adam:Optimization}, which helps the learning by applying adaptive optimisation to the learning rate of the network. 
        \item The learning rate is an important hyperparameter of a deep learning model. It is usually the most crucial value during training and should be configured using trial and error.
        Depending on the size of your dataset, a specific learning  rate decay may be needed. The learning rate decay is a technique that allows the learning rate to fall during successive training epochs, until it converges.
        A high learning rate at the start prevents the network from memorising noisy data, whereas decaying the learning rate improves complex pattern learning.
       \item  Implement early stopping and monitor the learning process by looking at the training loss plot to prevent overfitting.
       \item  Add complexity to your model gradually, e.g. add more layers or use  off-the-shelf \ac{CNN} models, and obtain a performance improvement over time.
    \end{itemize}

\subsection{Testing the model}
When the model is trained, its accuracy and performance should be tested using the test subset of the training dataset. A test set can also be independent to the training dataset to evaluate the model performance. The main point to remember is that the test set should not have been used for the training or evaluation of the model, at all. 

The model's performance should be measured by computing appropriate metrics suitable to the task at hand. A list of most common metrics used in testing fish monitoring models are given in Tabel \ref{table:symb}. For classification tasks, Classification Accuracy (CA), Precision and Recall rates are appropriate metrics, while F1-score, which is a combination of precision and recall, can provide a better measure of model performance and is used in fish counting and localization tasks as shown in Tables \ref{table:cont} and \ref{table:loc}. The Intersection-Over-Union (IoU) is the appropriate metric for segmentation tasks, while the mean average precision (mAP) metric suits pixel-wise localization of fish in images.
Looking at Tables \ref{table:cont} to \ref{table:seg}, other metrics such as Mean Square Error (MSE) and Root MSE (RMSE) have also been used in the marine fish monitoring literature. These can be considered and used if required.

\subsection{Fine Tuning the model}
The performance and accuracy of the model could be improved if needed. The amount of this improvement is, though, strongly influenced by its current accuracy. 
This step may quickly become complicated, since increasing the model accuracy might require several steps such as adjusting the learning rate, collecting new data, or fully modifying the model's architecture.
You should keep this fine tuning step to a reasonable level. Otherwise, the model might overfit the data.

\subsection{Deploying the model}
Finally, the model deployment mode should be chosen. This depends on the application and the deployment requirements.
The model can be deployed to run on a local or remote device (on a web server, a docker container, a virtual private server (VPS), etc). This will determine whether the results can be accessed remotely or only within the local network.
It is recommended to use a cross-platform deployment method to avoid issues such as input/output data format, or the type of files used for storing data.

The most commonly used cross-platform model deployment method is Docker  \citep{Potdar2020,Abdul2019}, which is a virtualization software that allows setting-up and running other software environments on top of a base Linux distribution without the need to set-up virtual machines. Docker helps build, configure, and run applications using the same Docker file. Typically, Docker is the recommended approach for web applications. In this method, you can use Docker container or Docker host on your development machine. Docker container may be the easiest option for web applications.
You can also deploy your network to a remote machine via Docker. The advantage of using a container is that you can share the development environment and run tests of your model using multiple docker containers.
You can also install the Docker tool on your local machine to manage containers, so it is convenient.

\begin{figure*}[!ht]
\includegraphics[width=\textwidth]{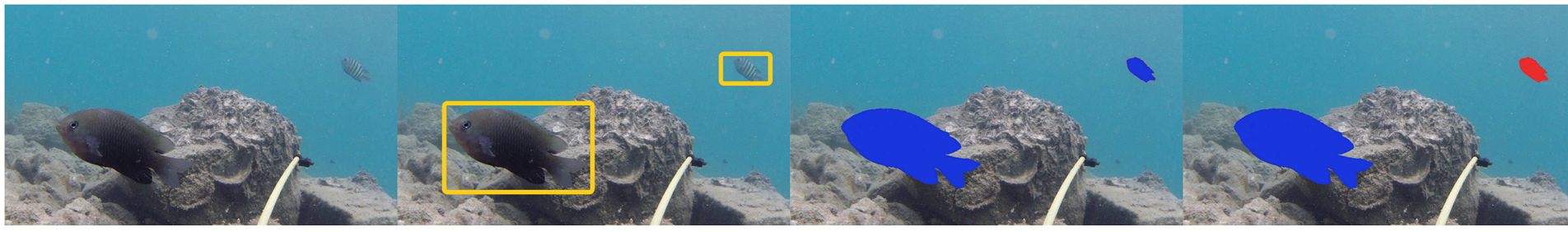}
\caption{Illustration of four typical fish monitoring tasks. From left: Fish Classification (\textit{i.e.} is there a fish in the image, or what type (class) of fish is in the image?); Fish Detection/Localization/Counting; Fish Semantic Segmentation, and Fish Instance Segmentation. } \label{CV_types}
\end{figure*} 



\begin{table*}

\centering
\caption{Performance metrics used to compare various surveyed works.}
\label{table:symb}
\arrayrulecolor[rgb]{0.647,0.647,0.647}
\resizebox{\linewidth}{!}{%
\begin{tabular}{>{\hspace{0pt}}m{0.211\linewidth}>{\hspace{0pt}}m{0.073\linewidth}>{\hspace{0pt}}m{0.652\linewidth}} 
\arrayrulecolor{black}\hline
Performance Metric & Symbol Used & Description \\ 
\arrayrulecolor[rgb]{0.647,0.647,0.647}\hline
\rowcolor[rgb]{0.929,0.929,0.929} Classification Accuracy & CA & The percentage of correct predictions. For
  multi-class classification, CA is averaged among all the classes. $CA = (TP + TN) / (TP + TN + FP + FN)$ \\
Precision & P & The fraction of true positives ($TP$), to the sum of $TP$ and false positives ($FP$). $P = TP/(TP+FP)$ \\
\rowcolor[rgb]{0.929,0.929,0.929} Recall & R & The fraction of true positives (TP) to the sum of TP and false negatives (FN). $R = TP/(TP+FN)$ \\
F1 score & F1 & The harmonic mean of precision and recall.  $ F1 = 2~
  \times (P \times   R)/(P + R)$ \\
\rowcolor[rgb]{0.929,0.929,0.929} Mean Square Error & MSE & Mean of the square of the errors between
  predicted and observed values \\
Root Mean Square Error & RMSE & 
Is the square root of the mean of the square of all of the errors.  \\
\rowcolor[rgb]{0.929,0.929,0.929} Mean Relative Error & MRE & The mean error between predicted and observed
  values,~in percentage \\
L2 error & L2 & Root of the squares of the sums of the
  differences~between predicted counts and the actual counts \\
\rowcolor[rgb]{0.929,0.929,0.929} Intersection over Union & IoU & A metric that evaluates how similar the predicted
  bounding box is to the ground truth bounding box.~ by dividing the area of overlap between the
  predicted and the ground truth boxes, by the area of their union. \\
The maximum number & MaxN & MaxN, the maximum number of the target
  species in any one frame. \\
\rowcolor[rgb]{0.929,0.929,0.929} Mean average precision & mAP & Depending on the detection difficulty, the mean
  $AP$ across all classes and/or total $IoU$ thresholds are used. \\
  Classification Error & CE & Is how often is the classifier incorrect and  also known as "Misclassification Rate". $CE = (FP + FN) / (TP + TN + FP + FN)$\\
\hline
\end{tabular}
}
\arrayrulecolor{black}


\end{table*}

\section{Applications of Deep Learning in Underwater Fish Monitoring}\label{secapps}

Deep learning has been widely used in marine environments with applications spanning from deep-sea mineral exploration  \citep{Juliani2021} to automatic vessel detection  \citep{Chen2019}. However, we confine the scope of this paper to only marine fish image processing, which typically includes four tasks of classification, counting, localization, and segmentation of underwater fish images, as shown in Fig. \ref{CV_types}.

Here, the goal is to assist the reader in understanding the similarities and differences across these tasks and their relevant \ac{DL} models and techniques. We provide a background of what each task involves, what previous works have been published toward addressing it using deep learning, and synthesize the literature on each task.

\subsection{Classification}
\label{seccls}
As its name infers, in visual processing, classification is the task of classifying images into different categories. There can be only two categories, i.e. a binary classification, in which the images are classified into two  groups, e.g. "fish" and "no fish", depending on the presence or absence of fish in an image (e.g. Deepfish dataset described in the first row of Table \ref{table:dataset}). The classification can also involve multiple "classes" or groups. For instance, consider assigning different underwater fish images into different groups based on the species (e.g. FishPak dataset in Table \ref{table:dataset}) present in them. 

Consider a manual procedure, in which images in a dataset are compared and relative ones are classified based on similar features, but without necessarily knowing what you are searching for in advance. 
This is a difficult assignment as there could be thousands of images in the dataset. 
Moreover, many image classification tasks involve images of different objects. 
It rapidly becomes clear that an automatic system, such as a \ac{DNN}, is required to complete this task quickly and efficiently.

Classification is the most widely-used and -studied underwater image processing task using \ac{DL}. In a previous work, we have covered the use of \acp{DNN} specifically for the task of underwater fish classification. We refer the reader to  \citep{saleh2022a} for a comprehensive review of prior art on classification.

\subsection{Counting}
\label{seccnt}

The purpose of the counting task is to predict the number of objects existing in an image or video.
Object counting is a key part of the workflow in many major \ac{CV} applications, such as traffic monitoring  \citep{Khazukov2020Real-timeParameters,Zhang_2017_CVPR}.
In the context of marine application and fish monitoring, counting may be used to map distinct species and monitor
fish populations for effective conservation. With the use of commercially available underwater cameras, data gathering can be done more comprehensively. 
It is, however, difficult to correctly count fish in  underwater habitats. To perform effective counting, models must understand the diversity of the items in terms of posture, shape, dimension, and features, which makes them complex. Meanwhile, manual counting is very time-consuming, costly, and prone to human error.

\ac{DL} affords a faster, less expensive, and more accurate alternative to the manual data processing methods currently employed to monitor and analyse fish counts. Table \ref{table:cont} lists several of the recent \ac{DL} techniques used for fish counting. Saleh \textit{et al.}  \citep{Saleh2020} created a novel large-scale dataset of  fish from 20 underwater habitats. They used \acp{FCN} for several monitoring tasks including fish counting and reported a \ac{MAE} of $0.38\%$.
\ac{DL} has the potential to be a more accurate method for assessing fish abundance than humans, with results that are stable and transferable between survey locations.
Ditria \textit{et al.}  \citep{Ditria2021a,Ditria2020b,Ditria2020c} compared the accuracy and speed of \ac{DL} algorithms for estimating fish population in underwater pictures and video recordings to human counterparts in order to test their efficacy and usability.
In single image test datasets, a \ac{DL} method performed $7.1\%$ better than human marine specialists and $13.4\%$ better than citizen scientists. For video datasets, \ac{DL} was better by $1.5\%$ and $7.8\%$ compared to marine and citizen scientists, respectively.

\begin{table*}
\begin{sideways}
\begin{minipage}{\textheight}

\centering
\caption{Summary of recent DL research works performing the task of fish counting}
\label{table:cont}
\arrayrulecolor[rgb]{0.647,0.647,0.647}
\resizebox{\linewidth}{!}{%
\begin{tabular}{>{\hspace{0pt}}p{0.18\linewidth}>{\hspace{0pt}}p{0.05\linewidth}>{\hspace{0pt}}p{0.05\linewidth}>{\hspace{0pt}}p{0.16\linewidth}>{\hspace{0pt}}p{0.16\linewidth}>{\hspace{0pt}}p{0.100\linewidth}>{\hspace{0pt}}p{0.048\linewidth}>{\hspace{0pt}}p{0.069\linewidth}>{\hspace{0pt}}p{0.186\linewidth}}
\arrayrulecolor{black}\hline
\multicolumn{1}{>{\centering\hspace{0pt}}m{0.18\linewidth}}{\textbf{Article }} & \multicolumn{1}{>{\centering\hspace{0pt}}m{0.05\linewidth}}{\textbf{DL Model}} & \multicolumn{1}{>{\centering\hspace{0pt}}m{0.05\linewidth}}{\textbf{Framework}} & \multicolumn{1}{>{\centering\hspace{0pt}}m{0.16\linewidth}}{\textbf{Data}} & \multicolumn{1}{>{\centering\hspace{0pt}}m{0.16\linewidth}}{\textbf{Annotation/Pre-processing/Augmentation}} & \multicolumn{1}{>{\centering\hspace{0pt}}m{0.100\linewidth}}{\textbf{Classes and Labels}} & \multicolumn{1}{>{\centering\hspace{0pt}}m{0.048\linewidth}}{\textbf{Perf. Metric}} & \multicolumn{1}{>{\centering\hspace{0pt}}m{0.069\linewidth}}{\textbf{Metric Value}} & \multicolumn{1}{>{\centering\arraybackslash\hspace{0pt}}m{0.186\linewidth}}{\textbf{Comparisons with other methods}} \\
\arrayrulecolor[rgb]{0.647,0.647,0.647}\hline

\rowcolor[rgb]{0.929,0.929,0.929} A realistic fish-habitat dataset to evaluate algorithms for underwater visual analysis \cite{Saleh2020}
 & ResNet-50 CNN & Pytorch & Authors-created database containing
  39,766 images for  20 habitats from remote coastal marine
  environments of tropical Australia and split to sub-dataset for four computer vision tasks:
  classification, counting, localization,
  and segmentation. & Each image was annotated by point-level and semantic segmentation labels & 20 classes of 20 different  fish habitat. & MAE & 0.38 & NA \\
  
  Annotated Video Footage for Automated Identification and
  Counting of Fish in Unconstrained Seagrass Habitats \cite{Ditria2021a} & ResNet-50 CNN & Pytorch & The dataset consists of 4,281 images and 9,429 annotations (9,304
  luderick, 125 bream) at the standard high resolution (1920 x 1080 p). &  Each image was annotated by drawing a bounding box and segmentation mask & 2 classes of fish & F1 & 92\% & NA \\
  
\rowcolor[rgb]{0.929,0.929,0.929} Automating the Analysis of Fish Abundance Using Object   Detection Optimizing Animal Ecology With Deep Learning \cite{Ditria2020b} & Mask R-CNN
  ResNet50 & Pytorch &  
  Authors-created database containing
   6,080 fish images  from   20 habitats from Tweed River Estuary in
  southeast Queensland 
    & Each image was annotated by segmentation mask & 1 class of fish & F1 & Image (95.4\%)   Video   (86.8\%) & The computer's performance in determining abundance was
  7.1\% better than human marine experts and 13.4\% better than citizen
  scientists in single image test datasets, and 1.5\% and 7.8\% higher in video
  datasets, respectively. \\

Deep learning for automated analysis of fish abundance:
the benefits of training across multiple habitats \cite{Ditria2020c} & Mask R-CNN
  ResNet50 & Pytorch & Authors created five datasets, each consisting of  4700 annotated luderick,
  total of 23500 images & Each image was annotated by drawing a Polygonal segmentation masks  around the region of
  interest (ROI) & 1 fish class & F1 & 87–92\% & NA \\
  
\rowcolor[rgb]{0.929,0.929,0.929} Deep learning with self-supervision and uncertainty
  regularization to count fish in underwater images \cite{Tarling2021DEEPIMAGES} & ResNet50
  CNN & Tensorflow & Authors created a data set of 500 labelled sonar images
  from video sequences & Each image was annotated by dot annotation & 3 classes of fish according to number of fish & MAE & 0.30\% & Comparison between DeepFish dataset 0.38\% and authors' benchmark
  result and their model 0.30\%. \\

Counting Fish and Dolphins in Sonar Images Using Deep
  Learning \cite{Schneider2020CountingLearning} & CNN & NA & Authors created a data set of 143 labeled sonar images from the Amazon
  River & Each image was annotated by counting number of fishes & 35 classes for fish and 4 for dolphin & MSE & Fish 2.11\%
  Dolphins 0.133\% & Comparing four Network Architectures, 
  DenseNet201, InceptionNetV2, Xception, and MobileNetV2 \\
  
\rowcolor[rgb]{0.929,0.929,0.929} Counting Fish in Sonar Images \cite{Liu2018a} & CNN & NA & Authors created a dataset of 537 labelled sonar images
  from video sequences & Each image was annotated by dot annotation & 1 class of fish & RMSE & 16.48\% & Comparison with other state-of-the-art approaches \\

Assessing fish abundance from underwater video
using deep neural networks \cite{Mandal2018c} & Faster R-CNN & Caffe & Authors created a dataset of 4909 labelled   images
  from video sequences & Each image was annotated by drawing a bounding box & 50 classes from 50 Different
  fish habitat. & mAP & 82.4\% & NA \\

\hline
\end{tabular}
}
\arrayrulecolor{black}

\end{minipage}
\end{sideways}
\end{table*}

Despite this high potential, \ac{DL} has not been thoroughly investigated for counting underwater fish.
One possible reason for the lack of comprehensive research for fish counting is the scarcity of large publicly available underwater fish datasets. In addition, properly annotating fish datasets to train robust \ac{DL} models is time-prohibitive and expensive.
Although the underwater fish counting is limited in the literature, several previous works have advanced the field in this area. For instance, 
Tarling \textit{et al.}  \citep{Tarling2021DEEPIMAGES} created a novel dataset of sonar video footage of mullet fish labelled manually with point annotations 
and developed a density-based \ac{DL} model to count fish from sonar images.
They counted fish by using a regression method   \citep{Xue2016} and achieved a \ac{MAE} of $0.30\%$.
Other researchers  \citep{Schneider2020CountingLearning,Liu2018a} used sonar images as well because they present substantially different visual characteristics compared to natural images.
Counting fish in sonar images, however, is substantially different
from counting fish in underwater video surveillance  \citep{Mandal2018c}. Unlike natural images, sonar images present unique visual characteristics and are in lower resolution due to the specific imaging forming principle.

Using \ac{DL}, a computer can be taught to identify fish in underwater images, thus eliminating the subjectivity of humans in counting fish. 
However, its use for fish population and count analysis is dependent on the model performance on a set of well-defined performance metrics and parameters, which is in itself a challenge. In section \ref{sectech}, we discussed how one can train high-performance \ac{DL} models, how the use of the current \ac{DL} pipeline (and other methodologies) can be improved, and how future \ac{DL} models can be designed for better assessing fish population including their abundance and their location, which is the subject of the next subsection.

\subsection{Localization}
Object localization is an essential task in \ac{CV}, where the goal is to locate all instances of specified objects  (e.g. fish, aquatic plants and coral reef) in images.
Marine scientists assess the relative abundance of fish species in their environments regularly and track population variations.
Various \ac{CV}-based fish sample methods in underwater videos have been offered as an alternative to this tedious manual assessment.
Though, there is no perfect method for automated fish localization. 
This is mostly owing to the difficulties that underwater videos bring, such as illumination fluctuations, fish movements, vibrant backgrounds, shape deformations, and variety of fish species.

To address these issues, several research works have been carried out, which are listed in Table \ref{table:loc}. Saleh \textit{et al.}  \citep{Saleh2020} have developed a fully convolutional neural network that performs localizing of fish in realistic fish-habitat images with a high accuracy. 
Jalal \textit{et al.}  \citep{Jalal2020} introduced a hybrid method  based on  motion-based feature extraction that combines optical flow   \citep{Beauchemin1995TheFlow} and Gaussian mixture models   \citep{MOG2006} with the YOLO deep learning technique  \citep{Chaudhari2020} to identify and categorise fish in unconstrained underwater videos using temporal information.
They achieved fish detection F-scores of $95.47\%$ and $91.2\%$ on  LifeCLEF 2015 benchmark   \citep{LifeCLEF2014} and their own dataset, respectively.
Gaussian mixture is an unsupervised generative modelling approach that may be used to learn first and second order statistical estimates of input data features  \citep{MOG2006}.
Within an overall population, this is used to indicate Normally Distributed subpopulations.
The weakness of Gaussian mixture is when trained on videos with some fish but no pure background, the fish are modelled as background as well, resulting in misdetections in subsequent video frames  \citep{salman2019automatic}.
In order to compensate for the Gaussian mixture's weakness, optical flow can be used to extract features which are solely caused by underwater video motion.
The pattern of apparent motion of objects, surfaces, and edges in a visual scene generated by the relative motion of an observer and a scene is known as optic flow  \citep{Beauchemin1995TheFlow}.

Knausgard \textit{et al.}  \citep{Knausgard2021a} also implemented  YOLO  \citep{Chaudhari2020} for fish localization.
To overcome their small training samples, they employed transfer learning (explained in the next Section). The YOLO  technique achieved \ac{MAP} of $86.96\%$ on the Fish4Knowledge dataset  \citep{Giordano2016}. 
YOLO-based object detection systems have been also used in several other research to robustly localize and count fish  \citep{Jalal2020,Xu2018UnderwaterApplications,Knausgard2021a}.
To test how well Yolo could generalise to new datasets,  \citep{Xu2018UnderwaterApplications} used it to localize fish in underwater video using three very different datasets. The model was trained using examples from only two of the datasets and then tested on examples from all three datasets. 
However, the resulting model could not recognise fish in the dataset that was not part of the training set.

Other \ac{CNN} models have also been adapted to robustly detect fish  under a variety of benthic background and illumination conditions. For instance,  \citep{Villon2016a} and  \citep{ChoiFish2015} used GoogLeNet \citep{GoogLeNet2015}, while  \citep{Labao2019b} used an ensemble of Region-based Convolutional Neural Networks  \citep{Ren2015FasterNetworks} that are linked in a cascade structure by Long Short-Term Memory networks  \citep{LSTM}.  
In addition, Inception  \citep{InceptionV3} and ResNet-50  \citep{He2015ResNet} were examined in  \citep{Zhuang2017} for fish detection and recognition based on weakly-labelled images. Furthermore,  \citep{Han2020a} and  \citep{Li2015b} used Fast R-CNN (Region-based Convolutional Neural Network)  \citep{Ren2015FasterNetworks} to detect and count fish.


Table \ref{table:loc} demonstrates that state-of-the-art methods (e.g. YOLO and Fast R-CNN) can achieve high accuracy in localization tasks. These methods generally train object detectors from a wide 
variety of training images  \citep{Felzenszwalb2010,Girshick2014} in a fully supervised manner. 
The drawback is that these models depend on instance-level annotations, e.g. tight bounding boxes need to be drawn around fish in training datasets. This is  time-consuming and labour-intensive and make the use of \ac{DL} in marine research very challenging, if not impossible. 
In Section \ref{sec_wsl} we discuss how this critical issue can be addressed using weakly supervised localization of objects, where only binary image-level labels showing the existence or absence of an object type are needed for training.

Similar to fish classification, counting, and localization, fish segmentation, i.e. detecting the entire body of fish in an image is a critical task in marine research and applications. In the next subsection, we discuss how \ac{DL} can be used to perform fish segmentation and how it is useful in marine research.

\begin{table*}
\begin{sideways}
\begin{minipage}{\textheight}

\centering
\caption{Summary of recent DL research works performing the task of fish localization}
\label{table:loc}
\arrayrulecolor[rgb]{0.647,0.647,0.647}
\resizebox{\linewidth}{!}{%
\begin{tabular}{>{\hspace{0pt}}p{0.19\linewidth}>{\hspace{0pt}}p{0.05\linewidth}>{\hspace{0pt}}p{0.05\linewidth}>{\hspace{0pt}}p{0.16\linewidth}>{\hspace{0pt}}p{0.16\linewidth}>{\hspace{0pt}}p{0.100\linewidth}>{\hspace{0pt}}p{0.048\linewidth}>{\hspace{0pt}}p{0.069\linewidth}>{\hspace{0pt}}p{0.176\linewidth}}
\arrayrulecolor{black}\hline
\multicolumn{1}{>{\centering\hspace{0pt}}m{0.19\linewidth}}{\textbf{Article }} & \multicolumn{1}{>{\centering\hspace{0pt}}m{0.05\linewidth}}{\textbf{DL Model}} & \multicolumn{1}{>{\centering\hspace{0pt}}m{0.05\linewidth}}{\textbf{Framework}} & \multicolumn{1}{>{\centering\hspace{0pt}}m{0.16\linewidth}}{\textbf{Data}} & \multicolumn{1}{>{\centering\hspace{0pt}}m{0.16\linewidth}}{\textbf{Annotation/Pre-processing/Augmentation}} & \multicolumn{1}{>{\centering\hspace{0pt}}m{0.100\linewidth}}{\textbf{Classes and Labels}} & \multicolumn{1}{>{\centering\hspace{0pt}}m{0.048\linewidth}}{\textbf{Perf. Metric}} & \multicolumn{1}{>{\centering\hspace{0pt}}m{0.069\linewidth}}{\textbf{Metric Value}} & \multicolumn{1}{>{\centering\arraybackslash\hspace{0pt}}m{0.176\linewidth}}{\textbf{Comparisons with other methods}} \\
\arrayrulecolor[rgb]{0.647,0.647,0.647}\hline

  \rowcolor[rgb]{0.929,0.929,0.929} Marine Animal Detection and Recognition with Advanced
  Deep Learning Models  \citep{Zhuang2017} & ResNet-10 CNN & NA & The dataset is made of 73 videos from the public datasets   Fish4Knowledge
    & Each image was annotated by drawing a bounding box & 1 class of fish & F1 & 0.07\% & NA \\

  Fish detection and species classification in underwater
  environments using deep learning with temporal information  \citep{Jalal2020} & Yolo - CNN & TensorFlow & The dataset is made of two  datasets
  93 videos from LifeCLEF 2015 fish dataset  And
  an authors-created database containing 4418 videos
    & Each image was annotated by drawing a bounding box and species name & 15 classes of 15 different fish species. & F1 & LCF-15 95.47\%
  UWA 91.2\% & Comparison with other state-of-the-art approaches \\
  
 \rowcolor[rgb]{0.929,0.929,0.929}Automatic fish detection in underwater videos by a deep
  neural network-based hybrid motion learning system  \citep{salman2019automatic} & ResNet-152 CNN & TensorFlow & The dataset is made of 110 videos from two public datasets
  Fish4Knowledge and 
  LifeCLEF 2015 fish dataset & Each image was annotated by drawing a bounding box & 15 classes of 15 different fish species. & F1 & 87.44\% and 80.02\% respectively & NA \\

Temperate fish detection and classification: a deep learning based approach  \citep{Knausgard2021a} & YoloV3  - CNN & Pytorch & total of 27230 images catalogued into 23 different species from the   public datasets   Fish4Knowledge
    &  Each image was annotated by drawing a  bounding box & 23 classes  of 23 different fish species. & mAP & 86.96\% & NA \\
  
  \rowcolor[rgb]{0.929,0.929,0.929} Underwater Fish Detection Using Deep Learning for Water
  Power Applications  \citep{Xu2018UnderwaterApplications} & YoloV3  - CNN & Keras - TensorFlow & Authors-created database of underwater video sequences
  for a total of  70000 train/test frame &  Each image was annotated by drawing a  bounding box & 3 classes of fish & mAP & 54.74\% & NA \\

 Coral Reef Fish Detection and Recognition in Underwater
  Videos by Supervised Machine Learning: Comparison Between Deep Learning and
  HOG+SVM Methods  \citep{Villon2016a} & GoogLeNet CNN & NA & Authors-created database containing
   13000 fish thumbnails from videos &  Each image was annotated by drawing a  bounding box & 11 classes  of 8 different fish species. & F1 & 98 \% & Compare HOG+SVM
  With Deep Learning \\
  
  \rowcolor[rgb]{0.929,0.929,0.929} Fish identification in underwater video with deep convolutional neural network: SNUMedinfo at LifeCLEF fish task 2015  \citep{ChoiFish2015} & GoogLeNet CNN & NA & 20 videos from LifeCLEF 2015 fish dataset
    &  Each image was annotated by drawing a  bounding box & 15 classes  of 15 different fish species. & AP & 81\% & NA \\
  
 Cascaded deep   network systems with linked ensemble components for underwater fish detection   in the wild  \citep{Labao2019b} & RNN- LSTM &  NA & Authors-created database containing 18 underwater video
  sequences for a total of 327 train/test frame &  Each image was annotated by drawing a  bounding box and species name & 1 class of fish & F1 & 67.76\% & Comparison with R-CNN Baseline \\

\rowcolor[rgb]{0.929,0.929,0.929} A realistic fish-habitat dataset to evaluate algorithms for underwater visual analysis  \citep{Saleh2020}
 & ResNet-50 CNN & Pytorch & Authors-created database containing
  39,766 images for  20 habitats from remote coastal marine
  environments of tropical Australia and split to sub-dataset for 
  classification, counting, localization,
  and segmentation. & Each image was annotated by point-level and semantic segmentation labels & 20 classes of 20 different  fish habitat. & MAE & 0.38 & NA \\
  
 Marine Organism Detection and Classification from
  Underwater Vision Based on the Deep CNN Method  \citep{Han2020a} & VGG16 -RCNN & NA & The dataset is obtained from the video provided by the Underwater
  Robot Picking Contest, test set contains 8800 images. &  Each image was annotated by drawing a  bounding box & 3 classes of fish& mAP & 91.2\% & NA \\

\hline
\end{tabular}
}
\arrayrulecolor{black}

\end{minipage}
\end{sideways}
\end{table*}

\subsection{Segmentation} 

Semantic segmentation task is to predict a label from a set of pre-defined object classes for each pixel in an image  \citep{Shelhamer2017}. In the context of marine research, fish segmentation provides a visual representation of fish contour, which might be helpful for human expert visual verification or to estimate fish size and weight. Table \ref{table:seg} lists a number of research addressing the task of fish segmentation.

Saleh \textit{et al.}  \citep{Saleh2020} developed a \ac{FCN} model that performs fish Segmentation in realistic fish-habitat images with a high accuracy. 
Labao \textit{et al}.  \citep{Labao2019c} proposed a \ac{DL} model that can simultaneously localize fish, estimate bounding boxes around them and segment them using a unified multi-task \ac{CNN} in underwater videos.
Unlike previous approaches  \citep{Qian2016,Wang2021a} that relied on motion information to identify fish body, their proposed method predicts fish object spatial coordinates and per-pixel segmentation using just video frames independent of motion information. 
Their suggested approach is more resilient to camera motions or jitters since it is not dependent on motion information, making it more suitable for processing underwater videos captured by Autonomous Underwater Vehicles (AUVs).
Region Proposal Networks (RPN)  \citep{Faster_R_CNN} have been also used for fish segmentation in underwater videos  \citep{Alshdaifat2020}.
RPN is a \ac{FCN} that generates boxes around identified objects and gives them confidence scores of belonging to a specific class, simultaneously.

\begin{table*}
\begin{sideways}
\begin{minipage}{\textheight}

\centering
\caption{Summary of recent DL research works performing the task of fish segmentation}
\label{table:seg}
\arrayrulecolor[rgb]{0.647,0.647,0.647}
\resizebox{\linewidth}{!}{%
\begin{tabular}{>{\hspace{0pt}}p{0.18\linewidth}>{\hspace{0pt}}p{0.05\linewidth}>{\hspace{0pt}}p{0.05\linewidth}>{\hspace{0pt}}p{0.16\linewidth}>{\hspace{0pt}}p{0.16\linewidth}>{\hspace{0pt}}p{0.100\linewidth}>{\hspace{0pt}}p{0.048\linewidth}>{\hspace{0pt}}p{0.069\linewidth}>{\hspace{0pt}}p{0.186\linewidth}}
\arrayrulecolor{black}\hline
\multicolumn{1}{>{\centering\hspace{0pt}}m{0.18\linewidth}}{\textbf{Article }} & \multicolumn{1}{>{\centering\hspace{0pt}}m{0.05\linewidth}}{\textbf{DL Model}} & \multicolumn{1}{>{\centering\hspace{0pt}}m{0.05\linewidth}}{\textbf{Framework}} & \multicolumn{1}{>{\centering\hspace{0pt}}m{0.16\linewidth}}{\textbf{Data}} & \multicolumn{1}{>{\centering\hspace{0pt}}m{0.16\linewidth}}{\textbf{Annotation/Pre-processing/Augmentation}} & \multicolumn{1}{>{\centering\hspace{0pt}}m{0.100\linewidth}}{\textbf{Classes and Labels}} & \multicolumn{1}{>{\centering\hspace{0pt}}m{0.048\linewidth}}{\textbf{Perf. Metric}} & \multicolumn{1}{>{\centering\hspace{0pt}}m{0.069\linewidth}}{\textbf{Metric Value}} & \multicolumn{1}{>{\centering\arraybackslash\hspace{0pt}}m{0.186\linewidth}}{\textbf{Comparisons with other methods}} \\
\arrayrulecolor[rgb]{0.647,0.647,0.647}\hline

\rowcolor[rgb]{0.929,0.929,0.929} A realistic
  fish-habitat dataset to evaluate algorithms for underwater visual analysis  \citep{Saleh2020}
    &  ResNet-50 CNN & Pytorch & Authors-created database
  containing
   39,766 images   from   20 habitats from
  remote coastal marine environments of tropical Australia   and split to sub-dataset for   classification, counting and localization,
  and   segmentation. & Each image was annotated by  point-level and semantic
  segmentation labels & 20 classes of 20 Different
  fish habitat. & mIoU & 0.93\%
    & NA \\

 Weakly supervised underwater fish segmentation using affinity LCFCN  \citep{Laradji2021} &  ResNet-CNN & Pytorch & Public DeepFish dataset  \citep{Saleh2020a} & Each image was annotated by   segmentation labels & 20 classes of 20 Different  fish habitat & mIoU & 0.749\% & NA \\

\rowcolor[rgb]{0.929,0.929,0.929} Simultaneous  Localization and Segmentation of Fish Objects Using Multi-task CNN and Dense
  CRF  \citep{Labao2019c} & ResNet-CNN & TensorFlow & Authors-created dataset containing
  1525 images 
  from ten 10 different sites in
  central Philippines & Each image was annotated by drawing a bounding box and segmentation
  labels & 1 class of fish & AP & 93.77\% & NA \\

Semantic Segmentation of   Underwater Imagery: Dataset and Benchmark  \citep{Islam2020} & VGG16 -CNN & Keras - TensorFlow & Authors-created dataset containing
   1525 images of 8 object categories & Each image was annotated by  segmentation labels & 8 classes of 8 different object
  categories. & mIoU & 84.14\% & NA \\
  
  \rowcolor[rgb]{0.929,0.929,0.929} DPANet: Dual Pooling-aggregated Attention Network for fish segmentation  \citep{Zhang2021} &ResNet-50 CNN & Pytorch, & Two public datasets DeepFish  \citep{Saleh2020a} and SUIM  \citep{Islam2020} & Each image was annotated by  segmentation labels & 20 classes: 20 Different
  fish habitat. & mIoU & 91.08\%, 85.39\% & Comparison with other state-of-the-art
approaches \\

  Weakly-Labelled semantic segmentation of fish objects in underwater videos using a deep residual network  \citep{Labao2017a} & ResNet-FCN & TensorFlow & Authors-created dataset containing   several underwater videos from six different sites in Verde Island
  Passage, Philippines. & Each image was annotated with weakly-labelled
  ground truth derived from a
  motion-based background subtraction (BGS) & 1 class of fish & AP & 65.91\% & NA \\

\rowcolor[rgb]{0.929,0.929,0.929} Improved deep learning framework
  for fish segmentation in underwater videos  \citep{Alshdaifat2020} &  ResNet-CNN & TensorFlow & Two datasets extracted from the
  Fish4Knowledge to produce 2000 frames & Each image was annotated by drawing a bounding box and segmentation
  labels & 15 classes  of 15 different fish
  species. & AP & 95.20\% & NA \\

\hline
\end{tabular}
}
\arrayrulecolor{black}

\end{minipage}
\end{sideways}
\end{table*}

Computational efficiency is essential in the autonomy
pipeline of visually-guided underwater robots. For this reason,  \citep{Islam2020} developed SUIM-Net, a fully-convolutional encoder-decoder model that balances the trade-off between performance and computational efficiency.
On the other hand, for higher performance,   \citep{Zhang2021}  proposed Dual Pooling-aggregated Attention Network (DPANet) to adaptively capture long-range dependencies through a computationally friendly manner to enhance feature representation and improve not only the segmentation performance, but also its computational resources and time.

All previously discussed models use fully-supervised methods that require a large amount of pixel-wise annotations, which is very time-consuming and expensive, because a human expert must segment and label, for example, each fish in an image. To overcome this serious issue, weakly-supervised semantic segmentation models are used. These models do not need to be trained with pixel-wise annotation  \citep{Rajchl2016}. However, due to a lower level of supervision, training weakly-supervised semantic segmentation models is often a more challenging task.
Applying weakly labelled ground truth derived from motion-based adaptive Mixture of Gaussians Background Subtraction,  \citep{Labao2017a}
managed to get an average precision of 65.91\%, and an average recall
of 83.99\%.
Recently, several other weakly-supervised methods have been introduced to overcome the cost of a large amount of pixel-wise annotations. These new methods include bounding boxes  \citep{khoreva2017simple,Dai2015},
scribbles  \citep{Lin2016}, points  \citep{Laradji2021,bearman2016s}, and
even image-level annotation  \citep{Pathak2015ConstrainedCN,Wang2018d,Ahn2018,Huang2018,Wei2018}.
Since weakly-supervised methods are integral to success of important \ac{DL}-based segmentation tasks, in Section \ref{seclimit}, we discuss them further. 

In the previous subsections, we discussed how \ac{DL} is useful in a number of key applications in fish habitat monitoring. In the following Section, we discuss the many challenges on the way of developing \ac{DL} models for such applications.

\section{Challenges in underwater fish monitoring}\label{secdchl}

.

Underwater fish monitoring presents a series of challenges for \ac{DL}, which have been the focus of many research works. In this section, we first introduce the major enviromental challenges faced when developing underwater fish monitoring models. We then show that one of the approaches to properly address these enviromental challenges is to use DL. However, DL training for fish monitoring has its own challenges, which will be discussed in details.  

\subsection{Environmental challenges}
In order to work in underwater environments, monitoring models must be able to recognize objects and scenes in complex, non-trivial backgrounds. This presents both a challenge in the development and training of these models and in robustly testing them. The main environmental challenges in underwater visual fish monitoring can be categorized as follows:

\begin{enumerate}
    \item The environment is noisy including very large lighting variation. An object viewed from a distance is much less bright than a close-up object. These problems become more acute when the background is not uniform.
    \item Underwater scenes are highly dynamic, i.e. the scene's content and objects change very quickly. The background can  change from being completely occluded to being visible and vice versa.
    \item Depth and distance perception can be incorrect due to refraction. This is more severe for short distances.
    \item Images are affected by water turbidity, light scattering, shading, and multiple scattering.
    \item The image data are frequently under-sampled due to  low-resolution cameras and power constraints underwater.
\end{enumerate}

One of the main approaches used in literature to address these challenges is for the monitoring models to use hand-crafted features  \citep{Rova2007,Hu2012b,Fouad2014,Huang2014,Chuang2016,Ogunlana2015,Hossain2016,Wang2017a,Islam2019}. Hand-crafted features are defined by a human to describe a fish image. 
For example, a low-level feature can be the histogram of a texture or a Gabor filter response. As a more complex and representative feature, a mid-level feature can be a Scale-Invariant Feature Transform (SIFT) \citep{Lindeberg2012}, or a Histogram of Oriented Gradient (HOG) \citep{Dalal2005}. However, human-defined features cannot be applied to other datasets, and the definition of a human-defined feature is a time-consuming task, which restricts real-time detection and requires manual effort. Moreover, hand-crafted features are limited by human experiences, which may contain noise and are difficult to design. For example, a SIFT descriptor doesn't work well with lighting changes and blur.

Therefore, a fish image is transformed into a feature space that a computer can understand. The feature space is often based on a combination of low-level image features (for example, colour distribution and gradient), and other features in the image such as edges, shapes, and textures. 
Models using hand-crafted features, however, do not perform well under varying environmental conditions, and the feature space cannot be easily or robustly created. Additionally, the features created are too low-level and cannot be easily used for processing images from different sources.

An alternative way to build prediction models capable of working in the presence of these significant environmental challenges is to use \acp{DNN}. However, training effective DNNs require resolving some other challenges, which we discuss in the below subsections. We also describe some of the approaches in literature addressing them. The reviewed approaches in addressing these common challenges can provide a quick reference for future researchers developing DL-based fish monitoring models.

\subsection{Model Generalisation} \label{secgen}

Improving the generalization abilities of DNNs is one of the most difficult tasks in \ac{DL}. Generalization refers to the gap between a model's performance on previously observed data (i.e training data) and data it has never seen before (i.e testing data). This is a fundamental problem, with implications for any applications using deep neural networks to process image data, videos, etc. This challenge is even more pronounced when more difficult tasks such as fish recognition in underwater environments. 

Generalization problem happens usually because during training the network over-fits to the training data. In other words, the weights of the network are adapted to produce a response that is best suited for reproducing the training examples. During testing, the network produces a response that is a compromise between the different training examples. This mismatch is a common cause of poor performance on test data, which is often referred to as a network over-fitting to the training data, even when the network has been trained for many epochs. The reason it occurs is that the network "memorizes" the training data during the training. 
The training data can become quite large, consisting of hundreds of thousands or millions of examples. This makes the issue of network over-fitting quite significant. In the last few years, there has been significant research efforts toward solving the problem of over-fitting to improve model generalization. 

Previous works have shown that it is possible to prevent the network from over-fitting using techniques called regularisation  \citep{kukavcka2017regularization}. There are also some theoretical techniques to make the network more robust to training data. Below, we provide a brief overview of some of these techniques and how they have been applied to solve the problem of deep network over-fitting to training data, to improve generalisation in \ac{DL}.

\begin{itemize}

    \item \textit{Regularisation Term}: It is hypothesised that neural networks with fewer weight matrices can result in simpler models with the same capability as the complete model. A regularisation term is, therefore, added to the model loss function to remove some of the weight matrices components. The most popular methods of regularisation are L1 and L2. 
    For example,  Tarling \textit{et al.}  \citep{Tarling2021DEEPIMAGES}  showed that incorporating uncertainty regularisation improves performance of their multi-task network with ResNet-50  \citep{He2015ResNet} backend to count fish in underwater images.
    
    \item \textit{Batch normalisation}: Introduced in Section \ref{sec:cnn} as part of the convolutional layer in \acp{CNN}, batch normalisation 
    was first introduced by Ioffe and Szegedy  \citep{batch_norm} to decrease the effect of internal covariate shift.  Internal covariate shift is the shift in the mean and covariance of inputs and network parameters across a batch of examples. Internal covariate shift can impede the training of deep neural networks.
    Batch normalisation is used in almost any \ac{DL} model training, to improve the model generalisation. In the fish monitoring domain, for instance, Islam \textit{et al.} \citep{Islam2020} proposed an optional residual
    skip block consisting of three convolutional layers with batch
    normalisation and ReLU non-linearity after each convolutional
    layer to perform effective semantic segmentation of underwater imagery.
    
    \item \textit{Dropout}: Introduced in Section \ref{sec:cnn} as a common operation in \acp{CNN}, dropout 
    reduces the network dependency to a small selection of neurons and encourages more useful and robust properties and features of the dataset to be learnt. 
    When working with a complex neural network structure, dropout is frequently recommended to introduce additional randomisation, which helps with the generalisation capability of the network.
    For example,  Iqpal \textit{et al.}  \citep{Iqbal2021a} claimed that the inclusion of dropout layer has enhanced the overall performance of their proposed model for automatic fish classification.

\end{itemize}

\subsection{Dataset Limitation} \label{seclimit}

Preparing training datasets is one of the central and most time-consuming bottlenecks in developing \ac{DL} models, which require a large amount of data, e.g. a variety of underwater fish images in different environmental conditions, which should also be labelled and analyzed by humans for supervised learning. Due to these requirements, making a large dataset is most of the time, very challenging, which makes the datasets limited and small. However, When compared with \ac{DL} models trained with a large dataset, the convergence speed and training accuracy of the models trained with small datasets are much lower. 
Generally, increasing the size of training datasets by adding more data to them is the classic way to accelerate the training and improved accuracy of \ac{DL} models, but it is expensive. Therefore, in recent years, researchers have tackled the dataset limitation challenge by devising new ways described below.

\subsubsection{Data Augmentation} 

Data augmentation is a technique to increase the number of labelled examples required for \ac{DL} training. It artificially enlarges the original training dataset by introducing various transformations such as translation, rotation, scaling, and even noise, to the original data instances, to make new instances.
It is particularly relevant to the challenge posed when  the  quantity or quality of labelled data is insufficient  to train a \ac{DL} model. At the same time, data augmentation can be used to reduce the probability of overfitting and increase model generalisability. 
In contrast to the techniques listed above for improving model generalisation, data Augmentation addresses overfitting from the source of the problem (\textit{i.e.} the original dataset). 
This is done under the notion that augmentations can extract additional information from the original dataset by artificially increasing the size of the training dataset. 
It is also critical to consider data augmentation's "safety" (\textit{i.e.} the possibility of misleading the network post-transformation). For example, rotation and horizontal flipping are typically safe data augmentation techniques for fish classification tasks  \citep{Saleh2020,Sarigul2017c} but not safe on digit classification tasks, due to the similarities between 6 and 9.
A data augmentation technique is to use the super-resolution reconstruction method  \citep{Ledig2017} based on Generative Adversarial Network (GAN)  \citep{Goodfellow2014GenerativeAN} to enlarge the dataset with high-quality images. This has been previously used to improve small-scale fine-grained fish classification  \citep{Qiu2018}, and to increase models predictive performance (i.e. ability to generalise to new data)  \citep{Konovalov2019a} for underwater fish detection and automatic fish classification  \citep{Chen2018c}.

Using augmentation techniques such as cropping, flipping, colour changes, and random erasing together can result in enormously inflated dataset sizes. For example, Islam \textit{et al.}  \citep{Islam2020} used rotation, width shift, height shift, shear, zoom and horizontal flip for semantic segmentation of underwater imagery to significantly increase their dataset size.
Another data augmentation technique used during
training \ac{DL} models is scale jittering, which has been used in  \citep{Mandal2018c} for assessing fish abundance in underwater videos. Gaussian filtering to blur images and different degrees of rotation for fish recognition in underwater-drone with a panoramic camera is another augmentation technique used in the marine monitoring domain  \citep{Meng2018a}.

However, augmentation is not always favourable, as it might lead to large overfitting in cases with very few data samples. 
As a result, it is critical to determine the best subset of augmentation techniques to train your \ac{DL} model using a limited dataset.

\subsubsection{Transfer Learning}
Transfer Learning is preserving information obtained while solving one problem, and transferring the learned knowledge to another similar problem. For instance, one may initially train a network on a large object dataset, such as ImageNet that includes 1000 different object classes, and then utilise the learned network parameters from that training as the initial learning parameters in a new classification task, e.g. fish classification. 
In most cases, just the weights in convolutional layers are transferred, rather than the complete network, including fully connected layers. 
This is extremely useful since many image datasets have low-level spatial features and properties that are better learnt in massive datasets. For example, Zurowietz \emph{et al.}  \citep{Zurowietz2020} presented unsupervised knowledge transfer to use their limited amount of training data in order to avoid time-consuming annotation for object detection in marine environmental monitoring and exploration.

\subsubsection{Hybrid Features}
\ac{DL} architectures have demonstrated excellent capabilities in capturing semantic knowledge that is latent in image features.
Handcrafted features, on the other hand, can provide specific physical descriptions if they are carefully chosen. In addition, attributes of natural images have been demonstrated to be described differently by \ac{CNN} features and hand-crafted features. 
This means a feature's discriminative ability may behave differently on different datasets. 
Therefore, these two types of features may complement each other for better learning. 

However, increasing feature dimensions by fusing hand-crafted and DL-generated features can result in increased computational requirement. One way to avoid this is to initially utilise \ac{DL} features for a particular dataset, and later add hybrid features to enhance the performance.
As a result, when working with difficult datasets, such as uncommon and rare marine species, more sophisticated algorithms and techniques based on hybrid features may be required. 
In fact, several research groups have used such strategies to improve the performance of marine species recognition tasks. 

For instance,
Mahmood \textit{et al.} \citep{Mahmood2016} used  texture- and colour-based hand-crafted features extracted from their CNN training data to complement generic CNN-extracted features and achieved a classification accuracy higher than when using only generic CNN features when classifying corals.
A combination of CNN and hand-designed features have also been used in  \citep{Cao2016} for marine animal classification, again showing that their method achieves higher accuracy than applying CNN alone. In another work, Blanchet \textit{et al.} showed that aggregation of multiple features outperforms models using single feature-extraction techniques, for automated coral annotation in natural scenes  \citep{Blanchet2016}.

\subsubsection{Weakly-Supervised Learning} \label{sec_wsl}

\ac{DL} methods  \citep{YannLeCunYoshuaBengio2015} have consistently achieved state-of-the-art results in a variety of applications, specifically in fully supervised learning tasks like classification and regression  \citep{Li2009,Lin2014}. Fully supervised learning methods  create predictive algorithms by learning from a vast amount of training patterns, where each pattern has a label showing its ground-truth output  \citep{Kotsiantis2007}. Although the current fully supervised methods have been very successful in certain activities  \citep{DeVos2017,Worz2006,Mader2018}, they come with a caveat of requiring a large portion of the data to be labelled, and it is sometimes difficult or extremely time consuming to obtain ground-truth labels for the dataset. Thus, it is desirable to develop learning algorithms that are able to work with less labelled data (\textit{i.e.} weakly supervised)  \citep{Zhou2018,Oquab2015}.

Weak supervision in particular can be very useful in underwater fish monitoring, where the limited dataset size and the time- and cost-prohibitive nature of labelling limits achieving a useful dataset for developing effective, smart, and automated habitat monitoring tools and techniques. A number of works in literature have already used weak supervision for underwater fish habitat monitoring. For example, Laradji \textit{et al.}  \citep{Laradji2020AffinitySupervision} proposed a segmentation model that can efficiently train on underwater fish images, not manually segmented for training, but only labeled with simple point-level supervision.
This work demonstrated that in the marine monitoring context, weakly-supervised learning can effectively improve the accuracy and speed of model development with limited dataset sizes and limited labelling budget.

\begin{figure}[!t]
\centering
\includegraphics[width=0.48\textwidth]{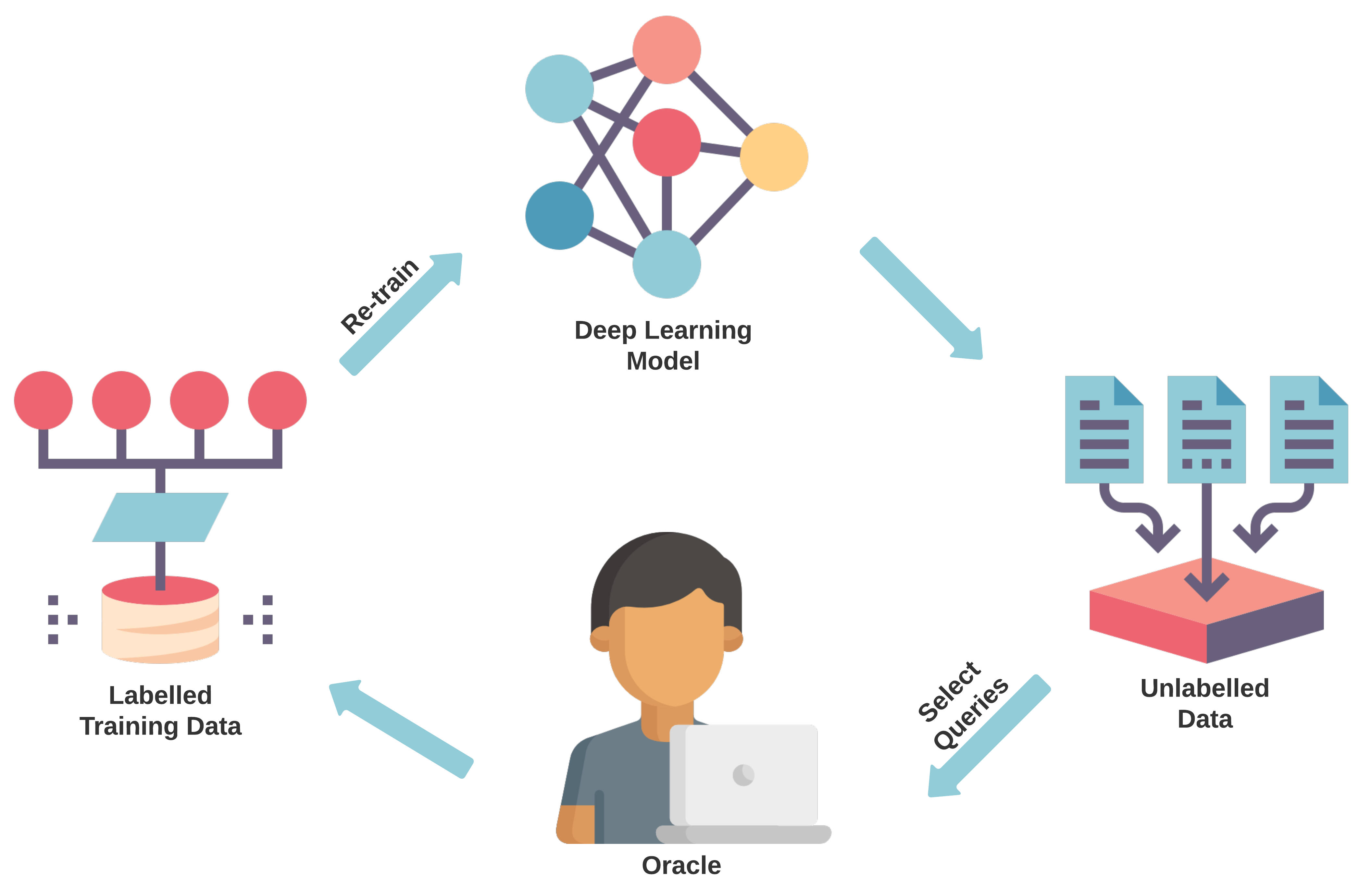}
\caption{Schematic diagram of Active Learning} \label{active}
\end{figure}

\subsubsection{Active Learning} 
Active learning is a sub-field of \ac{ML} and, more broadly, of \ac{AI}. 
In active learning, the proposed algorithm is allowed to be "inquisitive", that is, it is allowed to pick the data to learn, which in theory means the algorithm can do more with less guidance, similar to weak supervision. 
Active learning systems are seeking to solve the constraint of labelling by posing a questionnaire in the context of unlabeled examples to be labelled by an oracle (e.g. a human annotator). 
In this manner, the goal of the active learner is to attain high precision by 
using as few labelled examples as possible, thus minimising the expense of acquiring labelled data; see Figure \ref{active}. 

In many cases, the labels come for little or no cost, like the "spam" label that is used to mark spam emails, or the five-star rating that a user could post for a movie on a social networking platform. 
Learning methods use these labels and scores to help screen your spam email and recommend movies that you might enjoy. 
In these cases, certain labels are given free of charge, but for more sophisticated supervised learning tasks, such as when you need to segment a fish in an underwater environment, this is not the case.
For example, in  \citep{Nilssen2017} active learning has been used for the classification of species in underwater images from a fixed observatory. The authors proposed an active learning method that assigns taxonomic categories to single patches based on a set of human expert annotations, making use of cluster structures and relevance scores. This active learning method, compared to traditional sampling strategies, used significantly fewer manual labels to train a classifier.

\begin{figure*}[htbp]
\centering
\includegraphics[width=0.99\textwidth]{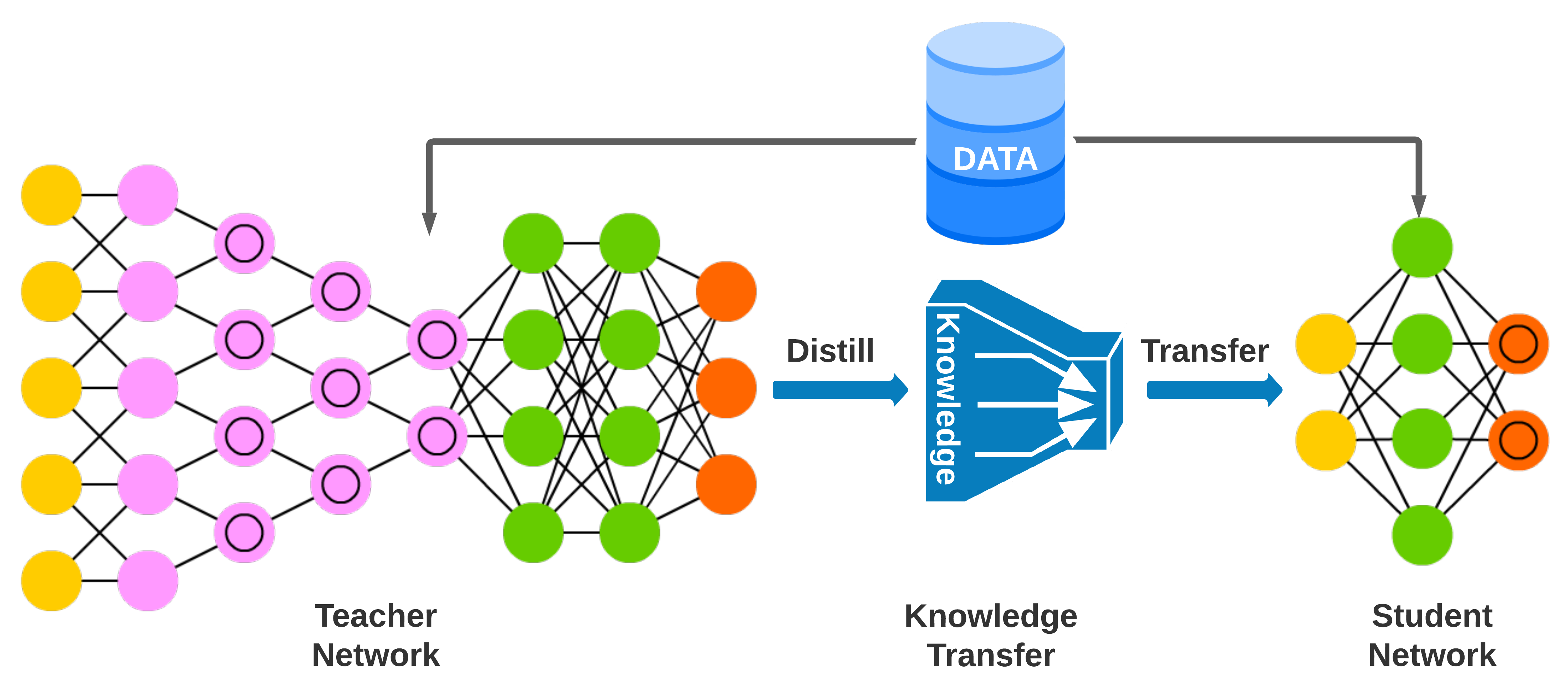}
\caption{Schematic diagram of knowledge distillation}
\label{fig:distillation}
\end{figure*}

\section{Opportunities in applications of DL to underwater fish monitoring}\label{secoppt}

New methodologies and strategies should be developed to advance \ac{DL} models for various underwater visual monitoring applications, including fish monitoring, and to bring them closer to their terrestrial monitoring equivalents. In a previous study that was focused on the task of fish classification  \citep{saleh2022a}, we have discussed some of the future research opportunities including (i) utilizing spatio-temporal data to add space and time domain information to the current training algorithms that mainly learn fish images regardless of their spatial and/or temporal correlation; (ii) Developing efficient and compact DL models that can be deployed underwater for real-time parsing of the fish images at the collection edge; (iii) Combining image data from multiple collection platforms for improved multi-faceted learning; and (iv) Automated fish measurement and monitoring from underwater captured images. Below, we expand on some of the previously discussed opportunities in  \citep{saleh2022a} and explore a few other prospective research areas for increasing the performance and usability of visual fish monitoring tasks.

\subsection{Knowledge Distillation for Underwater Embedded and Edge Processing}

DL models used for fish monitoring applications are usually very large containing millions of parameters and requiring extensive computational power. To deploy these models on resource-limited devices and in resource-constrained environments such as undersea monitoring sites, different hardware-emabled compression techniques such as quantizing and binarizing DNN parameters  \citep{Lammie2019} can be used, as discussed in  \citep{saleh2022a}.  
Another method that has seen a lot of interest and attention for compressing large-scale \ac{DL} models is knowledge distillation. 

Knowledge distillation is a technique for training a student (\textit{i.e.} a small network) to emulate a teacher (\textit{i.e.} ensemble of networks), as shown in Figure \ref{fig:distillation}. 
The primary assumption is that in order to achieve a competitive or even superior performance, the student model should imitate the teacher model.
The main issue is, however, transferring the knowledge from a large teacher to a smaller student.
To that end, Bucilua\textit{ et al.}  \citep{Bucila2006} proposed model compression as a way to transfer knowledge from a large model into a small model without sacrificing accuracy.
In addition, several other model compression approaches have been developed, and the community has shown an increasing interest in knowledge distillation, due to its potentials  \citep{Amadori2019,Wang2020,Rassadin2017,Kushawaha2021}.

A significant research opportunity lies in applying Knowledge distillation into embedded devices and underwater video processors to achieve online and more effective surveillance with high accuracy while using limited resources. This is particularly useful because of the limitations of transferring data from underwater sensors and cameras, and due to the challenging underwater communication in the Internet of Underwater Things~ \citep{Jahanbakht2021}.

\subsection{Merging Image Data from Multiple Sources}

As discussed in  \citep{saleh2022a}, to train more effective DNNs, multiple data collection platforms like Autonomous Underwater Vehicles (AUVs) or inhabited submarines can give varied visual data from the same monitoring subject. This can provide additional monitoring information, such as fish distribution patterns. 
Although it is straightforward to combine multiple data sources for training a DL network, several issues should be addressed in future research. These include possible preprocessing on part of data to make it compatible with the rest of the training dataset, class-wise weights (i.e. when you have an imbalanced dataset), and the number of outputs of a network. In addition, multiple training data sources, in particular, when using AUVs or  submarines, incurs significant data collection and manual labelling cost, which is not always viable. 

For this reason, some researchers have focused on learning from data with the least amount of human-labeling. To reduce human-labelled data cost, several methods have been proposed to train models on data that are unlabeled  \citep{Shimada2021} or only have pseudo-labels  \citep{Wu2018}. Future research can advance this further by developing faster and cheaper annotating tools for underwater fish images. 

\subsection{Automatic Fish Phenotyping From Underwater Images}

Automatic fish phenotyping, i.e. extracting their weight, size, and length, in their natural habitats can provide invaluable information in better understanding marine echosystems and fish ecology  \citep{Goodwin2021}. Although many studies have addressed fish monitoring in aquaculture and fish farm settings  \citep{Li2021a,Zhao2021}, monitoring fish for measurement in natural habitats remain mostly unexplored, and can be investigated in future research. These research should address problems such as low visibility and light, fish occlusion and overlap, which are shared with aquculture monitoring. However, other problems unique to natural habitats such as cluttered background environments and underwater distance measurement should be addressed too.

\subsection{Visual Monitoring of Fish Behavior and Movements}
Although some telemetry and satellite tracking devices can be used in limited settings  \citep{Lennox2017}, fish monitoring in their natural habitats over a period of time is not achievable using these techniques mainly due to the hostile underwater signal communication medium  \citep{Jahanbakht2021}. For instance for tracking fish movements, schooling, and behavior, new visual monitoring techniques should be devised. 
A possible direction for future studies is to devise better understanding of fish vision characteristics  \citep{Boudhane2016} and their implications in the current and next generation of automated DL-based tracking systems  \citep{li2021tracking} and marine object detection  \citep{Moniruzzaman2017}. An example of an alternative tracking method is presented in  \citep{Zhao2019}, where the image-based identification and tracking method for fish is designed based on biological water quality monitoring. To improve the fish tracking task, some techniques can also be combined with visual image enhancement algorithms. For instance, when the image enhancement methods are used, the underwater images can be corrected for distortion and noise, and the fish tracking task can be easily performed. In  \citep{Saberioon2016}, the authors studied the potential of underwater fish monitoring by using visual and underwater sensing methods.

Another challenging research area is developing novel underwater fish tracking algorithms, using DL or other technologies, with low power consumption and real-time speed. For this, various hardware technologies and techniques used in other domains such as biomedical applications~ \citep{Azghadi2020HardwareApplications} can be explored.
Of course, any automated vision-based tracking system should be validated through real-world trials, which is a significant undertaking requiring many resources, in order to ensure the accurate and real-time tracking of fish.

\section{Summary and Conclusion}\label{seccncl}
The goal of this article was to provide researchers and practitioners a summary of the contemporary applications of \ac{DL} in underwater visual monitoring of fish, as well as to make it easier to apply \ac{DL} to tackle real challenges in fish-related marine science.

DL has progressed as a technology capable of providing unprecedented benefits to various aspects of marine research and fish habitat monitoring. We envision a future where DL, complemented by many other advances in monitoring hardware and underwater communication technologies  \citep{Jahanbakht2021}, is widely used in marine habitat monitoring for (1) data collection and feature extraction to improve the quality of automatic monitoring tools; and (2) to provide a reliable means of surveying fish habitats and understanding their dynamics. 
We expect that such a future will allow marine ecosystem researchers and practitioners to increase the efficiency of their monitoring efforts. To achieve this, we need concentrated and coordinated data collection, model development, and model deployment efforts. We also need transparent and reproducible research data and tools, which help us reach our target sooner.

\section*{Acknowledgement}
This research is supported by an Australian Research Training Program (RTP) Scholarship and Food Agility HDR Top-Up Scholarship. We also acknowledge the Australian Research Council for funding awarded under their Industrial Transformation Research Program.

	

\bibliographystyle{cas-model2-names}


\bibliography{references}

\end{document}